\documentclass{article}
\usepackage{iclr2027_conference,times}
\usepackage[T1]{fontenc}

\usepackage{amsmath,amsfonts,bm}

\def\eqref#1{equation~\ref{#1}}
\def\1{\bm{1}}

\DeclareMathAlphabet{\mathsfit}{\encodingdefault}{\sfdefault}{m}{sl}
\SetMathAlphabet{\mathsfit}{bold}{\encodingdefault}{\sfdefault}{bx}{n}

\usepackage{hyperref}
\usepackage{url}
\usepackage{xcolor}
\usepackage{booktabs}
\usepackage{graphicx}
\usepackage{needspace}

\makeatletter
\long\def\@makecaption#1#2{%
  \vskip\abovecaptionskip
  \sbox\@tempboxa{\small #1: #2}%
  \ifdim \wd\@tempboxa >\hsize
    {\small #1: #2\par}%
  \else
    \global \@minipagefalse
    \hb@xt@\hsize{\hfil\box\@tempboxa\hfil}%
  \fi
  \vskip\belowcaptionskip}
\makeatother

\title{Adapting Prior-Data Fitted Networks for\\
Tabular Anomaly Detection}

\author{Maximilian Bershtman \qquad Niv Cohen \\
{\normalfont Faculty of Electrical and Computer Engineering} \\
{\normalfont Technion -- Israel Institute of Technology} \\
{\normalfont Haifa, Israel} \\
{\normalfont\texttt{max.bershtman@gmail.com} \quad \texttt{cohen.niv@gmail.com}}}

\iclrfinalcopy
\hypersetup{
  hidelinks,
  pdftitle={Adapting Prior-Data Fitted Networks for Tabular Anomaly Detection},
  pdfauthor={Maximilian Bershtman and Niv Cohen},
  pdfsubject={Unsupervised tabular anomaly detection with ZEN and FOCUS}
}

\begin{document}

\maketitle
\lhead{Preprint}

\begin{abstract}
While deep features have transformed anomaly detection in images and video, their impact on tabular data has been less substantial, partly due to the limited availability of strong deep representations. Recently, prior-data fitted networks (PFNs) have emerged as a promising source of such representations for tabular data. In this work, we investigate how PFN representations can be adapted and leveraged for anomaly detection.

The question is harder than it looks. No anomalies are available before
deployment, so model parameters cannot be tuned with supervision, and
the reference set that defines normal behavior may itself contain the
very anomalies it is supposed to reveal. We begin our study using frozen TabPFN
features. Scoring each sample by its distance to its nearest neighbors
in feature space already gives strong results. We identify which
layers to use and a feature-extraction procedure suited to the task. Next, to 
further improve performance, we use the reference set to fine-tune the model, so that the
resulting features better separate normal samples from anomalies. On
the ADBench benchmark, our fine-tuning free approach (ZEN) reaches a
higher mean AUROC than every baseline, and our fine-tuned method
(FOCUS) improves on it further. Our approach also generalizes across
PFN models~\footnote{Code:
\href{https://github.com/Maximilianb1/zen-focus-tabular-anomaly-detection}{\nolinkurl{github.com/Maximilianb1/zen-focus-tabular-anomaly-detection}}}.
\end{abstract}

\vspace{-0pt}
\section{Introduction}
\label{sec:intro}
\vspace{-0pt}

Tabular data is the most common data format in real-world machine learning
\citep{hollmann2023tabpfn}. This paper studies anomaly detection on tabular
data, one of the oldest tasks in the field and still one of its most active
\citep{han2022adbench,yoon2026retabad,li2026ofatad}. The task is
unsupervised: a detector receives a reference set of samples that describe
normal behavior, and must score how strongly each new sample deviates from
them. High scores should single out samples such as fraudulent
transactions, unusual patient records, or compromised network traffic. The
detector faces two unknowns at once. First, the anomalies: they are
rare and their structure is unknown in advance, which is why the task cannot
be posed as supervised classification. Second, the reference set: it
may itself be contaminated, as when fraud investigators can verify only
a small fraction of transactions and the unverified rest enter the
historical record as normal \citep{dalpozzolo2018fraud}.

We therefore study two settings. Our main
setting is the \emph{contaminated regime}, where the reference set contains
anomalies at an unknown rate and nothing removes them before a detector
sees it. The \emph{clean regime}, where the reference set is anomaly-free
but unlabeled, is the complementary setting.

Existing detectors handle these unknowns only partly. Detectors that
excel on a clean reference set can degrade sharply once anomalies
enter it \citep{marszalek2026tactic}, because an anomaly already in
the reference set makes the test anomalies that resemble it look
normal. Classical detectors remain
remarkably strong \citep{liu2008isolation,zhao2019pyod}, but every
deployment starts from zero, with nothing carried over. Deep methods
learn richer representations
\citep{livernoche2024dte,yin2024mcm,li2025tccm,dai2025noise}, but they
too train one model per dataset, and their gains have not proven robust
across datasets. The ADBench benchmark study reached a sobering
conclusion: no unsupervised detector in its comparison came out reliably
better than the others, ``emphasizing the importance of algorithm
selection'' \citep{han2022adbench}.

Anomaly detection in images solved a similar problem through \emph{transfer}:
over the last decade it was rebuilt on representations reused from models
pretrained once at scale
\citep{yosinski2014transferable,bommasani2021foundation}. Features
taken from generically pretrained networks and scored by simple
$k$-nearest-neighbor distance outperformed elaborate self-supervised
detectors \citep{bergman2020dn2}, and the same kind of features now
underlies state-of-the-art industrial defect detection
\citep{cohen2020spade,roth2022patchcore}. Tabular data could not follow this path, lacking comparably strong generic representations
\citep{grinsztajn2022tree,shwartz2022tabular}. PANDA, an image anomaly detection method, names the absence of generic
extractors for some data modalities as the main limitation of its
approach \citep{reiss2021panda}.

Recently, prior-data fitted networks (PFNs), with TabPFN chief among
them, have emerged as strong pretrained models for tabular data, and
as a potential candidate feature extractor. These models are transformers
pretrained on synthetic tabular data
\citep{hollmann2023tabpfn,hollmann2025tabpfnv2,grinsztajn2026tabpfn3},
and they match or beat strong tuned baselines on supervised tabular
prediction \citep{hollmann2025tabpfnv2}. They work by in-context
learning: given a new table as context, they predict for query samples
in a single forward pass, with no dataset-specific training.

Here, we ask how such a model can best be leveraged for tabular anomaly detection. So far, PFN-based work on the task has taken two
routes. One pretrains a dedicated in-context detector on synthetic
anomaly-detection tasks
\citep{marszalek2026tactic,shen2025fomo0d,ding2026outformer}. The
other keeps TabPFN frozen and uses it as
a plug-in scorer over its predictions
\citep{tabpfnextensions,ding2026outformer}. Adapting a pretrained PFN
to the task itself remains mostly unexplored, although adaptation
of PFNs has shown some success for supervised learning \citep{feuer2024tunetables}. Notably, adaptation is especially interesting as a
detector built by adapting TabPFN would inherit future improvements of
the PFNs. Moreover, PFNs are never pretrained to find
anomalies, so adapting one also measures how much its representations
carry beyond their pretraining objective. This raises the question we
study: can a general-purpose tabular foundation model be adapted into a
competitive anomaly detector?

\begin{figure}[t]
\centering
% Source: ../Figures drafts/fig1_mechanism_v26.svg (gen_fig1_v26.py), rendered by headless Chrome from fig1_print_v26.html.
\includegraphics[width=\textwidth]{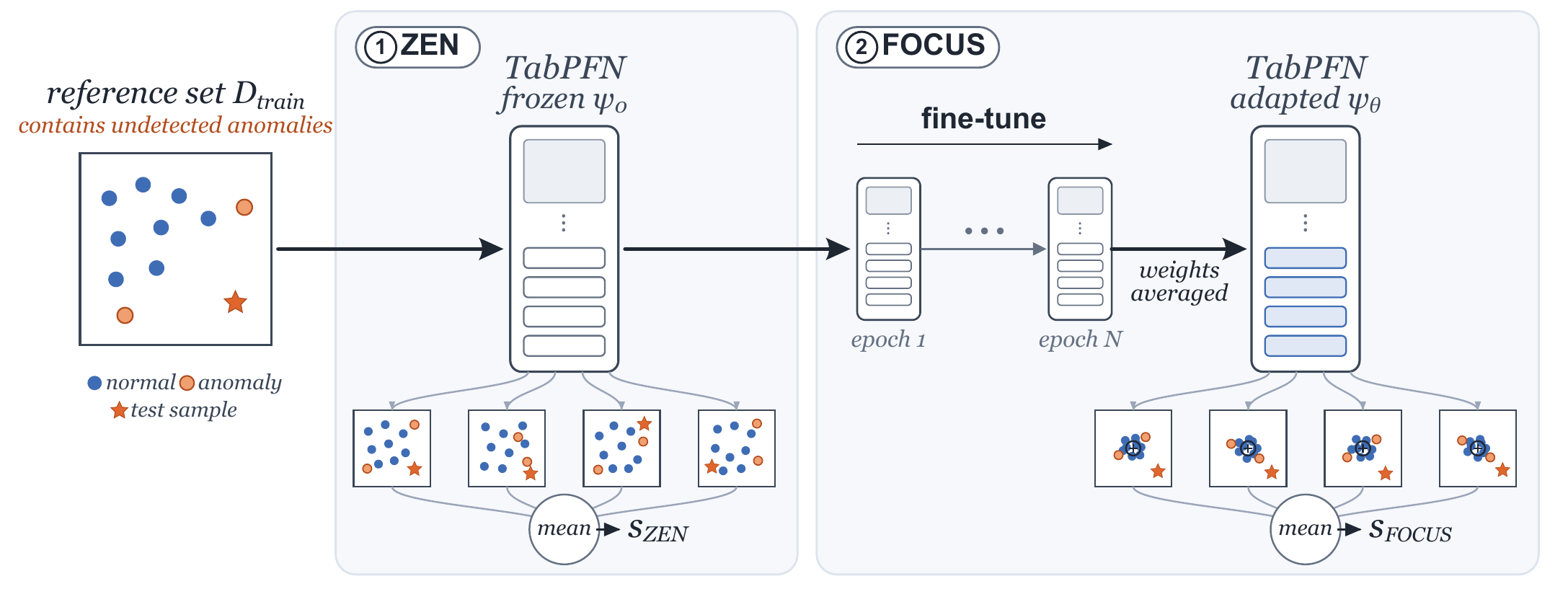}

\caption{Overview of ZEN and FOCUS. The reference set contains undetected
anomalies (orange). (1) ZEN embeds the reference set with frozen TabPFN,
using only the last four of its 24 blocks (bottom slabs). Each test
sample is scored by its distance to its nearest reference samples in
each block's embedding, and the four scores are averaged. (2) FOCUS
fine-tunes the model so that the reference set's embeddings contract
toward the fixed center $c$, and averages the weights of its epoch
checkpoints (two epochs) into one adapted model. The adapted
embeddings are scored as in (1). The drawing is simplified;
Secs.~\ref{sec:zen} and~\ref{sec:focus} give the full methods.}
\label{fig:fig1}
\end{figure}

This paper shows the answer is yes. We reach it in two steps that
adapt TabPFN, first without changing its weights and then by
updating them (Figure~\ref{fig:fig1}). First, ZEN (\textit{Zero-training
Embedding Neighbors}) scores a test sample by its distance to its
nearest neighbors in the frozen model's embedding space. We develop a
feature-extraction procedure suited to the task and weight each
reference sample by how much it can be trusted, so the score stays
reliable even when the reference set is contaminated. Second, FOCUS
(\textit{Fine-tuned One-Class Unsupervised Scoring}) fine-tunes the model so
that its features better separate the normal data from anomalies.
%Following previous works such as DeepSVDD and PANDA \citep{ruff2018deepsvdd,reiss2021panda}.%,
%we do so by encouraging the model to map the reference set toward a
%fixed center, with each reference sample weighted by how much it can be
%trusted, as in ZEN.
 Our evaluation for the contaminated settings finds that ZEN already reaches a higher mean
AUROC than every baseline in our comparison, and FOCUS the highest of
all.

\needspace{12\baselineskip}
\paragraph{Contributions.}
\begin{itemize}
  \item We introduce ZEN, a training-free anomaly detector on frozen
    TabPFN embeddings whose score stays reliable even when the reference set
    is contaminated (Sec.~\ref{sec:zen}).
  \item We propose FOCUS, which builds on ZEN: a trust-weighted
    compactness fine-tune of TabPFN that reaches the highest mean AUROC
    in both regimes (Secs.~\ref{sec:focus} and~\ref{sec:results}).
  \item Our approach is not tied to a single PFN model: applied without
    any change to four other tabular foundation models, it improves
    each of them significantly 
    (Sec.~\ref{sec:analysis}).
\end{itemize}

\vspace{-0pt}
\section{Method}
\label{sec:method}
\vspace{-0pt}

\subsection{Preliminaries}
\label{sec:pfn}
\vspace{-0pt}

\paragraph{Prior-data fitted networks.}
A prior-data fitted network (PFN) is a transformer trained to perform
supervised learning in a single forward pass \citep{hollmann2023tabpfn}.
It receives a labeled dataset
$D_{\mathrm{train}}=\{(x_i,y_i)\}_{i=1}^{n}$ as its context and a query
sample $x_{\mathrm{test}}$. From these it outputs a predictive
distribution $q_{\theta}(y\mid x_{\mathrm{test}},D_{\mathrm{train}})$,
taking no gradient step on $D_{\mathrm{train}}$. In Bayesian terms, that
forward pass approximates a posterior predictive distribution. A prior
over data-generating mechanisms $\phi\in\Phi$ induces
\begin{equation}
\label{eq:ppd}
p(y\mid x, D_{\mathrm{train}})\;\propto\;\int_{\Phi}
p(y\mid x,\phi)\,p(D_{\mathrm{train}}\mid\phi)\,p(\phi)\,d\phi.
\end{equation}
While this integral is intractable, a PFN approximates it: synthetic
datasets are sampled from the prior, and the network is trained to
predict each dataset's held-out labels from the rest of that dataset
(Appendix~\ref{app:config} states the objective). Pretraining happens
once; every new table afterwards is handled in context. TabPFN uses a
synthetic prior built on structural causal models, and it matches or
beats strong tuned baselines
\citep{hollmann2023tabpfn,hollmann2025tabpfnv2}. %Here we use TabPFN-3
%\citep{grinsztajn2026tabpfn3}.

\paragraph{TabPFN as a feature extractor.}
Beyond its predictions, TabPFN, like any transformer, computes an
internal representation of every sample it processes, which we use in
this work. Because it is an in-context model, every embedding depends
on the whole reference set. In addition, TabPFN embeds a sample differently depending on whether it is provided as context or used as a query. 
We therefore embed each reference sample as a
query against the rest of the set, leave-one-fold-out, following
\citet{ye2025closerlook}. Appendix~\ref{app:config} gives the details,
and the labels used for TabPFN's context.

\paragraph{Anomaly detection setting.}
\label{sec:problem}
The anomaly detection setting we study provides an unlabeled reference set
$D_{\mathrm{train}}=\{x_i\}_{i=1}^{n}\subset\mathbb{R}^{d}$, assumed
mostly normal. It is evaluated on a test set $D_{\mathrm{test}}$
composed of normal samples and anomalies. A detector assigns each test
sample a score $s(x)$, oriented such that higher means more anomalous.
Ground-truth labels are used only for evaluation. The two regimes we
investigate differ in one assumption about $D_{\mathrm{train}}$: in the
\emph{clean regime} it contains no anomalies, and in the
\emph{contaminated regime} it contains anomalies at an unknown rate.
Which regime holds is often unknown at deployment time.
\vspace{-0pt}

\subsection{ZEN: Zero-training Embedding Neighbors}
\label{sec:zen}
\vspace{-0pt}

Anomalies tend to lie where normal data is sparse. The classical
way to act on this is density-estimation based scoring: rate each test sample by how far
it sits from the reference samples. The usual measure is the distance to
the $k$ nearest reference samples; it is one of the oldest anomaly
detectors and, on clean tabular data, still one of the strongest
(Sec.~\ref{sec:results}). Our approach ZEN applies a similar rule, but in the
embedding space of the frozen TabPFN model. Embed every sample, then score each test
sample by its mean distance to the $k$ nearest reference samples:
\begin{equation}
\label{eq:knn}
s(x)\;=\;\frac{1}{k}\sum_{j\in N_k(x)}\bigl\lVert \psi(x_j)-\psi(x)\bigr\rVert_2,
\end{equation}
where $\psi(x)$ is a sample's embedding and $N_k(x)$ indexes the $k$
reference samples whose embeddings are nearest to $\psi(x)$. We use
$\psi_{0}^{\smash{(l)}}(x \mid C)$ for the frozen TabPFN's embedding of $x$
after transformer block $l$, given the context $C$; the index $0$ stands
for the original, frozen weights, as opposed to the fine-tuned weights
$\theta$ of Sec.~\ref{sec:focus}. TabPFN has 24 transformer blocks, and ZEN takes the
score from the last four of them, a window we denote $W$
(Appendix~\ref{app:window} ablates this choice). I.e., Eq.~\ref{eq:knn} is
computed on $\psi_{0}^{\smash{(l)}}$ for each block $l \in W$. Distances
in different blocks live on different scales, so each block's scores are
standardized before they are averaged. The mean and standard deviation
come from the reference samples' own leave-one-out distances. The
average over the four blocks is the score of one run of the model. ZEN
uses this score together with two techniques that handle a reference
set containing anomalies, explained next.

\paragraph{PFN feature extraction.}
We can get a more informative score by embedding the data several times rather
than once, each time with a different subset of the features. ZEN
therefore runs the frozen model on $B{+}1$ \emph{feature subsets}. $B$ of
them each hold a random half of the features,
$S_1,\dots,S_B\subset\{1,\dots,d\}$, and the last one, $S_{B+1}$, uses
all features together. Because the model uses its context to calculate embeddings, each feature
subset yields a different representation of the data. To regulate the
extracted features so that they sit on a similar scale, we also add to
the context of each $S_i$ synthetic samples drawn uniformly on
$[-1,1]$ (Appendix~\ref{app:config} gives every constant of the
extraction and Appendix~\ref{app:context} ablates them).

\paragraph{Soft cleaning.}
Under contamination the reference set contains anomalies, and an
anomalous test sample that lands next to one of them may receive a
small $k$NN distance and pass as normal. We therefore weight the
reference samples by how well their own neighbors support them. Let
$a_i$ be the mean distance of reference sample $i$ to its $k$ nearest
other reference samples. Each of the $B{+}1$ feature subsets and four
blocks yields one such distance. We standardize each over the reference
set and average the values. The
result is one isolation estimate $\hat a_i$ per reference sample, shared
by every feature subset and block. Sample $i$ receives the \emph{trust weight}
$w_i=\exp(-\lambda\,\hat a_i)$, and every distance to it is divided by
$w_i$:
\begin{equation}
\label{eq:soft}
\tilde s(x)\;=\;\frac{1}{k}\sum_{j\in N_k(x)}
\frac{\bigl\lVert\psi(x_j)-\psi(x)\bigr\rVert_2}{w_j},
\end{equation}
where $N_k(x)$ is the neighbor set of Eq.~\ref{eq:knn}, found with the
plain distance. A reference sample that sits far from its own
neighbors, as an undetected anomaly typically does, receives a small
weight: its distances are inflated, so a test sample next to it no
longer looks normal, while the sample itself is never removed. At
$\lambda{=}0$ every weight is one and the cleaning is off.
Appendix~\ref{app:sens} provides a sensitivity study.

\paragraph{The scoring function.}
For each feature subset $S_i$, Eq.~\ref{eq:soft} is computed per block
$l\in W$, standardized, and averaged over the four blocks, giving the
score $s_i(x)$. ZEN averages the $B$ random subsets and gives that
average the same weight as the weight of the full set of features, $S_{B+1}$:
\begin{equation}
\label{eq:zen}
s_{\mathrm{ZEN}}(x)\;=\;\frac{1}{B}\sum_{i=1}^{B}s_i(x)
\;+\;s_{B+1}(x).
\end{equation}

% P148c (Max): Sec. 2.3 opens p. 5; no top float on that page, so Table 1 ([t]) goes to the top of p. 6.

\suppressfloats[t]
\subsection{FOCUS: Fine-tuned One-Class Unsupervised Scoring}
\label{sec:focus}
\vspace{-0pt}

ZEN uses the embedding space exactly as pretraining left it. Yet, that
space is shaped for supervised prediction on synthetic tasks, not for
separating the normal samples of a given dataset from its anomalies.
Image anomaly detection faced the same situation and moved past its
frozen baselines by adapting pretrained features to the normal training
data \citep{reiss2021panda,reiss2023meanshifted,cohen2022transformaly}.
FOCUS brings this adaptation to TabPFN. Following the fixed-center
convention of DeepSVDD and PANDA \citep{ruff2018deepsvdd,reiss2021panda},
it fine-tunes the last blocks of the model such that the last layer embeddings of
the reference set contract toward a fixed center $c$, the trust-weighted
mean of the frozen model's final-block embeddings of that set. Under
contamination the reference set contains anomalies, so the trust
weights enter the loss: each reference sample pulls toward the center
in proportion to how much it can be trusted (according to the weights explained in \textit{Soft cleaning} above). The weights and center are
computed once from the frozen model and never updated. We write
$\psi_{\theta}$ for the fine-tuned extractor. Appendix~\ref{app:focusdetails}
gives the loss and every detail of the fine-tune, and
Appendix~\ref{app:ftabl} ablates its constants.

\paragraph{Averaging the fine-tuning epochs.}
Adapting pretrained features to the normal data can collapse them.
DeepSVDD avoids the collapse by constraining the network and fixing the
center \citep{ruff2018deepsvdd}. PANDA reports that naive adaptation
often deteriorates the features and counters it with early stopping
combined with a regularizer \citep{reiss2021panda}. Anomaly detection
provides no labeled validation set that would allow us to decide how to regularize the compactness finetuning in a supervised manner. Here we take a
simpler route: FOCUS trains for two epochs on every dataset and
averages the weights saved after the first and after the second epoch,
in the spirit of stochastic weight averaging \citep{izmailov2018swa}.
No further regularization is added.
Once adapted, the fine-tuned model replaces the frozen one in every
step of ZEN, using the same feature subsets, the same contexts, and the same
soft-cleaned readout. We call the
resulting method $s_{\mathrm{FOCUS}}$.

\begin{table}[t]
\centering
\caption{Main results, sorted by contaminated-regime AUROC. Each regime
has two columns: the mean AUROC ($\times 100$) over the 47 ADBench
datasets, and the mean rank over the regime's 18 methods. The best
value per column is in bold and the second best is underlined.
$^{\mathrm{d}}$ marks a deep learning method, trained per dataset;
$^{\mathrm{t}}$ marks a TabPFN-based method.}

\label{tab:main}
\smallskip
\scriptsize
\renewcommand{\arraystretch}{0.86}

\setlength{\tabcolsep}{5pt}
% Body: anon-release paper_assets/table1.tex (build_paper_assets.py on the F102 official per-dataset means).
\begin{tabular}{lrrrr@{\hspace{14pt}}lrrrr}
\toprule
Method & \multicolumn{2}{c}{Contaminated} & \multicolumn{2}{c}{Clean} & Method & \multicolumn{2}{c}{Contaminated} & \multicolumn{2}{c}{Clean}\\
\cmidrule(lr){2-3}\cmidrule(lr){4-5}\cmidrule(lr){7-8}\cmidrule(lr){9-10}
 & AUROC & rank & AUROC & rank & & AUROC & rank & AUROC & rank\\
\midrule

FOCUS$^{\mathrm{t}}$ (ours) & \textbf{80.0} & \textbf{5.9} & \textbf{85.9} & \underline{4.9} & PCA & 72.5 & 8.9 & 76.7 & 9.0\\
ZEN$^{\mathrm{t}}$ (ours) & \underline{79.7} & \underline{6.0} & \underline{85.2} & 5.6 & TCCM$^{\mathrm{d}}$ & 72.2 & 9.7 & 80.8 & 7.6\\
IForest & 76.4 & 7.2 & 79.6 & 8.4 & DeepSVDD$^{\mathrm{d}}$ & 70.9 & 11.4 & 74.7 & 11.8\\
TabPFN-PL$^{\mathrm{t}}$ & 74.6 & 8.8 & 78.8 & 9.4 & OCSVM & 70.1 & 11.4 & 75.6 & 10.8\\
COPOD & 74.4 & 8.4 & 74.5 & 11.3 & SOD & 69.1 & 11.7 & 58.7 & 15.6\\
HBOS & 74.4 & 8.6 & 77.4 & 10.0 & COF & 66.4 & 12.5 & 55.7 & 15.4\\
DTE$^{\mathrm{d}}$ & 74.3 & 9.8 & 81.5 & 7.9 & LODA & 66.0 & 12.7 & 69.8 & 12.9\\
raw $k$NN & 73.9 & 7.8 & 85.1 & \textbf{4.4} & LOF & 65.4 & 11.8 & 82.9 & 6.2\\
ECOD & 73.9 & 9.1 & 74.2 & 12.0 & 
CBLOF & 73.7 & 9.2 & 81.7 & 7.8 \\
\bottomrule
\end{tabular}
\end{table}

\vspace{-0pt}
\section{Results}
\label{sec:results}
\vspace{-0pt}

\subsection{Experimental setup}
\label{sec:setup}
\vspace{-0pt}

\paragraph{Data and protocol.}
We evaluate on all 47 classical tabular datasets of the ADBench benchmark
\citep{han2022adbench}. They span 80 to 619{,}326 samples and 3 to
1{,}555 features (Appendix~\ref{app:dstats}). In the clean regime the
training data (reference set) takes 70\%
of the normal samples and the test set holds the remaining normals and
every anomaly, according to the one-class convention
\citep{ruff2018deepsvdd,li2025tccm}. In the contaminated regime a
stratified 70/30 split preserves the anomaly rate on both sides, as in
ADBench. Datasets above 10{,}000 samples are first subsampled to
10{,}000, stratified and without replacement. We also note a leak in
the official pipeline, which duplicates the samples of the twelve
datasets with fewer than 1{,}000 samples before the split, so that most
test samples there have a twin in the training set. We fix that and evaluate these
datasets at their natural size instead (Appendix~\ref{app:leak}).
Features are scaled to $[-1,1]$ by a MinMax scaler fitted on the
training split, identically for every method. Seeds 0 through 4 were
fixed in advance, and each seed draws a fresh split.
Appendix~\ref{app:protocol} gives the protocol in full.

\paragraph{Baselines.}
The comparison covers the 13 classical and deep detectors that ADBench
itself benchmarks through PyOD \citep{han2022adbench,zhao2019pyod}. To these we add recent detectors
with public code: TCCM \citep{li2025tccm}, DTE
\citep{livernoche2024dte}, the official unsupervised TabPFN extension
\citep{tabpfnextensions}, and a TabPFN pseudo-labeling control (pseudo-labels are assigned using a PCA-based anomaly score). No
method, ours included, is tuned per dataset: baselines run with their
authors' recommended settings, ours with one fixed configuration per
regime. Every number is computed by us under this protocol;
Appendix~\ref{app:impl} gives every configuration.

\paragraph{Metrics.}
We report AUROC. AUPRC is reported in Appendix~\ref{app:auprc} and wall-clock
inference times in Appendix~\ref{app:runtime}. Per-dataset values are
means over the five seeds. Appendix~\ref{app:seedstd} reports the
spread. Each method is summarized by its mean AUROC and mean
rank across the 47 datasets, and every two methods are compared with a
Wilcoxon signed-rank test
\citep{demsar2006statistical}.

\subsection{Main results}

\begin{figure}[t]
\centering
% Source: final/figures/fig2_optA1_v10.pdf (build_results_v10.py; numbers in paper_results_stats_v10.json).
\includegraphics[width=\textwidth]{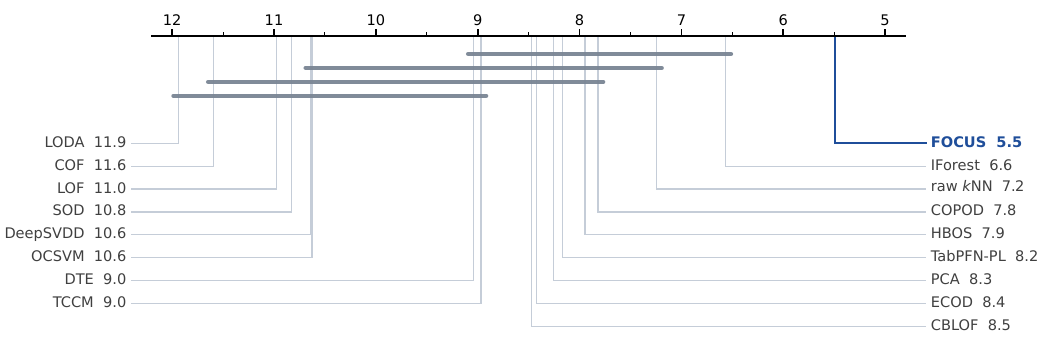}
\caption{Contaminated regime: mean rank of FOCUS and of every ranked
baseline over the 47 datasets, best rank on the right (we only report FOCUS here, so ranks differ from Table~\ref{tab:main}). FOCUS is tested
against each of the 16 baselines with a Wilcoxon signed-rank test,
Holm-corrected over the 16 comparisons. Bars among the baselines join
methods whose pairwise differences are not significant. FOCUS has a
statistically significant advantage in every comparison.
Appendix~\ref{app:blindcd} details the construction.}
\label{fig:main}
\end{figure}

\begin{figure}[t]
\centering
% Source: final/figures/fig3_v10.pdf (build_results_v10.py).
\includegraphics[width=\textwidth]{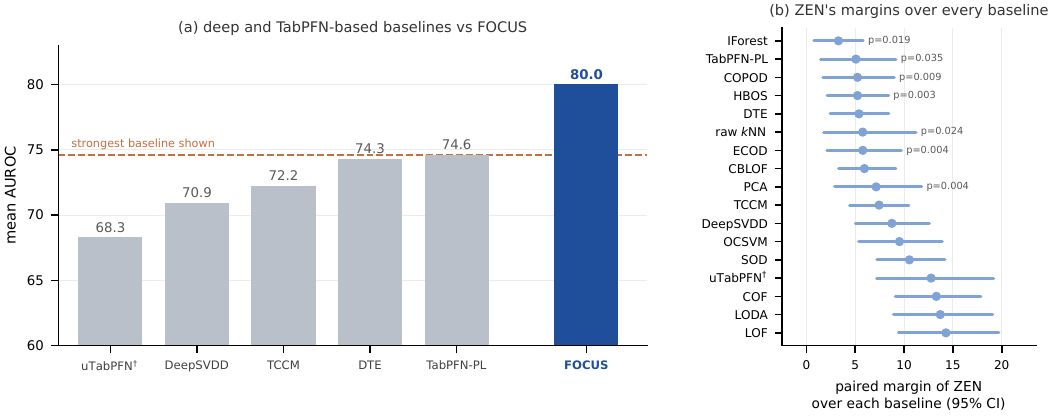}
\caption{Additional results, contaminated regime. (a) Mean AUROC
of the deep and TabPFN-based detectors and of FOCUS over the 47
datasets. Dashed line: the strongest baseline shown (TabPFN-PL).
(b) Paired AUROC margin of ZEN over each baseline. Dot: mean; bar:
95\% bootstrap interval over datasets. Every margin is positive and
significant (Wilcoxon $p<0.05$); the $p$-value is printed where it
exceeds $0.001$. $^{\dagger}$uTabPFN covers 44 of 47 datasets.}
\label{fig:support}
\end{figure}

\paragraph{Contaminated regime.}
Methods that perform very well in the clean regime often perform much
worse under contamination. Table~\ref{tab:main} shows the pattern in
each method's two regime columns. The nearest-neighbor detectors suffer
the most: raw-feature $k$NN loses 11 AUROC points, and LOF loses 17.
Isolation Forest degrades far less and becomes the strongest baseline.
Our methods sit above it. ZEN, our training-free version, reaches a
higher mean AUROC than every baseline, 3.3 points above Isolation
Forest ($p=0.02$). FOCUS improves on ZEN by a further 0.3 points and
reaches the highest mean AUROC of all methods, 3.6 points above
Isolation Forest ($p=0.01$) and 6.1 above raw-feature $k$NN
($p=0.02$). ZEN and FOCUS also hold the two best mean ranks. We
compare FOCUS with every ranked baseline at once. For each comparison we use the
Wilcoxon signed-rank test with a Holm correction, and find FOCUS
significantly ahead of all the baselines (Figure~\ref{fig:main}). Against
the deep and TabPFN-based detectors, the closest to ours in spirit,
FOCUS leads by a wide margin, and ZEN's margin over every baseline is
also positive and significant (Figure~\ref{fig:support}). Under AUPRC
(Table~\ref{tab:auprc}) FOCUS keeps the best mean in both regimes;
Appendix~\ref{app:perregime} adds per-regime critical-difference
diagrams and the complete pairwise matrix.

\paragraph{Clean regime.}
When the reference set contains no anomalies, the picture changes:
raw-feature $k$NN is the strongest baseline and sits well above every
other baseline (Table~\ref{tab:main}, clean columns). FOCUS and ZEN
reach the two highest mean AUROCs, but $k$NN keeps the best mean rank.
Our gains over the deep and TabPFN-based detectors remain large: FOCUS
leads every one of them by more than four AUROC points, with
statistical significance. On clean data the gain from fine-tuning is
significant: FOCUS improves on ZEN on 31 of the 47 datasets
($p=0.02$).

In practice, it is often unknown how contaminated the training set is,
if at all. A practitioner who chooses FOCUS
without knowing whether the clean or the contaminated regime holds
therefore matches the best classical detector if the reference set
turns out clean, and gains 3.6 AUROC points over the best baseline if
it does not.


\vspace{-0pt}
\section{Discussion}
\label{sec:analysis}
\vspace{-0pt}

\begin{figure}[t]
\centering
% Source: final/figures/paper_fig_analysis_v18.pdf (build_fig4_v18.py; v16 numbers, panel (b) label wording only; stats in paper_fig4_stats_v18.json).
\includegraphics[width=\textwidth]{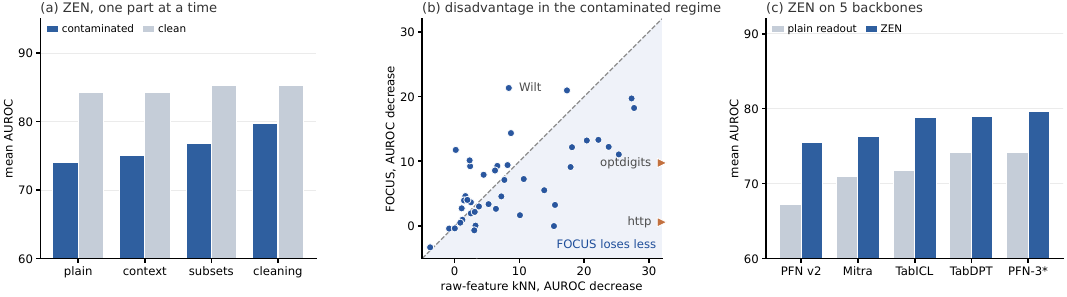}
\caption{Ablations and analysis. (a) The ablation ladder: mean AUROC of
the frozen model as ZEN's steps are added one at a time, the plain
readout, then the augmented context, the feature subsets and the soft
cleaning, in both regimes. (b) The disadvantage of moving from the
clean to the contaminated regime. the AUROC each method loses: FOCUS on the $y$-axis
against raw-feature $k$NN on the $x$-axis, one point per dataset. In the shaded region
FOCUS loses less due to contamination. Triangles are datasets beyond the axis. (c) ZEN's
first step, the plain readout,
and full ZEN on five backbones under contamination, every constant
unchanged. The asterisk marks the paper's backbone, TabPFN-3.}
\label{fig:analysis}
\end{figure}

\paragraph{How much does each step of the feature extraction help?}
We measure what each part of our method contributes, in the
contaminated regime. The starting point is the \emph{plain readout}:
the frozen model's embedding of the data, read by $k$NN distance and
nothing else (Sec.~\ref{sec:zen}). Figure~\ref{fig:analysis}a ablates
ZEN's three steps, one at a time (Appendix~\ref{app:ladder}
ablates every step in both regimes, with significance tests). The
plain readout reaches 74.1
AUROC, next to raw-feature $k$NN's 73.9 and the best deep detector.
The augmented context adds 0.9 points. The embedding with feature
subsets adds 1.9 points. The soft cleaning adds 2.8 points, the largest
step. Together the steps take the frozen model from 74.1 to 79.7, above
every baseline.

\paragraph{How and why is FOCUS resistant to contamination?}
Density-based detectors rate a sample by how far it sits from its
nearest reference samples. When the reference set holds undetected
anomalies, a test anomaly that is similar to them finds close neighbors
and appears normal. Figure~\ref{fig:analysis}b shows the cost the
contamination induces on each dataset: between the clean and the
contaminated regime raw-feature $k$NN loses 11 AUROC points on average,
while FOCUS loses only 5.9 points.

FOCUS limits the damage from contamination by finding the anomalies hiding in the
reference set and preventing them from ``hiding" the test anomalies. Its isolation
estimate scores every reference sample by how far it sits from its
nearest other reference samples. %The labels of the reference samples,
%which no method sees, let us check the soft cleaning afterwards. 
Used
as an anomaly score for the reference samples, the estimate reaches
76.7 AUROC (Appendix~\ref{app:ladder}, Figure~\ref{fig:estimate}). The
multi-subset averaging is essential here. It works because the runs use different features, and
therefore make different mistakes during soft cleaning. The trust weights built from the
pooled estimate average 0.70 on a reference anomaly and 1.16 on a
normal sample, so a test anomaly next to a reference anomaly no longer
looks normal. Appendix~\ref{app:ladder} gives the full breakdown.

\paragraph{Does our approach transfer to other tabular foundation models?}
Yes. We applied ZEN, the training-free part of our approach, with the
same feature subsets, blocks, constants and protocol, to four other
in-context tabular models, three from other groups. TabPFN v2
\citep{hollmann2025tabpfnv2}, TabICL \citep{qu2025tabicl} and Mitra
\citep{zhang2025mitra} are, like TabPFN-3, prior-data fitted networks
trained on synthetic priors; TabDPT \citep{ma2025tabdpt} is pretrained
on real tables instead. Figure~\ref{fig:analysis}c shows our improvements with these backbones
in the contaminated regime. Appendix~\ref{app:backbones} gives more details on both
regimes. ZEN lifts every backbone by roughly
five to eight AUROC points, and every gain is statistically
significant. The FOCUS fine-tune transfers as well: on TabPFN v2 it
lifts the plain readout by 3.0 AUROC points
(Appendix~\ref{app:backbones}). For two of these models, TabICL and
TabDPT, the results are about as strong as with TabPFN-3, our main
backbone.

\paragraph{How sensitive are ZEN and FOCUS to their constants?}
Not much: no constant sits at a sharp optimum. We ablated every
constant one at a time, on the same datasets, seeds and splits as the
main results. Under contamination, reading the last one to four blocks
of TabPFN gives a similar result, and reading more blocks, or earlier
ones, is worse. $k=50$ sits on a plateau (Table~\ref{tab:sens}). Every
$\lambda$ between 0.25 and 2 keeps ZEN above the strongest deep baseline.
Each added feature subset helps, with diminishing returns. For FOCUS, reducing the number of trainable blocks to three, for example, lowers mean AUROC by 0.5 points in both regimes. Training for one or
three epochs instead of two, or keeping the last epoch instead of the
average, changes it by at most 0.2 points.  Finally, the gain of the
soft cleaning grows with the contamination rate: when 2\% of the
reference set are injected anomalies the cleaning adds 0.1 AUROC
points, and when 20\% are, it adds 2.0. Appendix~\ref{app:stats} gives
the full details.

\vspace{-0pt}
\section{Related work}
\label{sec:related}
\vspace{-0pt}

\paragraph{Classical anomaly detectors.}
Classical tabular anomaly detection comes from several algorithm families
\citep{zhao2019pyod,han2022adbench}.
\emph{Neighborhood and local-density methods} score samples relative to nearby reference points.
kNN uses distance to neighboring samples, while LOF compares the local density of a sample with that of its neighbors
\citep{BreunigKNS00}.
\emph{Partitioning, projection, and ensemble methods} avoid explicit local-density estimation.
Isolation Forest isolates anomalies through recursive random partitions
\citep{liu2008isolation}, LODA aggregates weak detectors built on random one-dimensional projections
\citep{pevny2016loda}, and feature bagging combines detectors across random feature subsets
\citep{lazarevic2005feature}.
ZEN's feature subsets are closest to this last idea, though each subset changes the in-context representation itself.
\emph{Distributional and support-estimation methods} model the normal region more directly.
HBOS estimates feature-wise densities using histograms
\citep{goldstein2012histogram}, COPOD scores samples by their extremeness in the tails of empirical distributions
\citep{li2020copod}, and one-class SVM estimates the support of the data in a kernel feature space
\citep{scholkopf2001estimating}.

\paragraph{Deep anomaly detectors.}
An early self-supervised line of works
brought transformation-based objectives to tabular data: GOAD
classifies random affine transformations of the data
\citep{bergman2020goad}; NeuTraL AD learns the transformations
themselves \citep{qiu2021neutral}; ICL trains with an internal
contrastive objective \citep{shenkar2022icl}. DTE
estimates the diffusion time of a sample and scores it accordingly
\citep{livernoche2024dte}. TCCM learns a flow-matching-style
contraction field and scores by the deviation at one time step
\citep{li2025tccm}. MCM scores by masked cell modeling
\citep{yin2024mcm}. NPT-AD reconstructs masked features using a non-parametric transformer that alternates attention across features within each sample and attention across samples \citep{thimonier2024nptad}. Noise-evaluation methods learn to separate
the data from artificially corrupted copies of it \citep{dai2025noise}.
DeepSVDD introduced the fixed-center compactness objective FOCUS
builds on \citep{ruff2018deepsvdd}. All of these methods pay a per-dataset training cost; in our pipeline
the only training anywhere is FOCUS's two-epoch fine-tune.

\paragraph{Pretrained anomaly detectors.}
A recent line of work pretrains dedicated detectors once and applies
them to new datasets without further training: TACTIC
\citep{marszalek2026tactic} and FoMo-0D \citep{shen2025fomo0d} are
pretrained on synthetic anomaly-detection tasks for zero-shot
detection. OutFormer \citep{ding2026outformer} is pretrained on
synthetic labeled datasets and infers test labels in context.
OFA-TAD \citep{li2026ofatad} trains once on multiple source datasets
to generalize to unseen ones. These detectors commit to their anomaly
assumptions at pretraining time. Frozen TabPFN has also served as a
plug-in scorer: the official unsupervised extension scores anomalies
through the library's density construction over the frozen model
\citep{tabpfnextensions}, and OutFormer's appendix builds a similar
frozen TabPFN baseline \citep{ding2026outformer}. We differ from these prior attempts to use PFNs for anomaly detection:  ZEN uses the embeddings of a pretrained classification model
rather than its predictions, and FOCUS adapts them.

\paragraph{Transfer and adaptation in anomaly detection.}
In images, $k$NN on frozen pretrained features outperformed earlier
self-supervised detectors \citep{bergman2020dn2}. Anomaly localization methods were built on the same kind of features
\citep{cohen2020spade,roth2022patchcore}. PANDA, MSAD, and
Transformaly adapt the pretrained features to the normal training data
\citep{reiss2021panda,reiss2023meanshifted,cohen2022transformaly}.
Under contamination, SRR refines the training set with an ensemble of
one-class classifiers as training proceeds \citep{yoon2022srr}, LOE
infers latent labels for the unlabeled mix during training
\citep{qiu2022loe}, and EPHAD adjusts the outputs of a detector
trained on contaminated data with evidence gathered at test time
\citep{patra2025ephad}. FOCUS applies a related idea: it soft-cleans the reference set using the given initial representation. 

\vspace{-0pt}
\section{Limitations}
\vspace{-0pt}

\label{sec:limitations}
\paragraph{Hardware and runtime.}
Scoring with ZEN or FOCUS runs TabPFN forward passes for each of the
$B{+}1=6$ feature subsets. Both therefore need a GPU: on one A100, our method
processes 3,500 samples per minute 
(Appendix~\ref{app:runtime}). Although this is slower than many classical detectors, most of that time is taken by the forward passes
themselves, with ten leave-one-fold-out passes per feature subset
\citep{ye2025closerlook}. This cost can be cut in several ways. Fewer
leave-one-fold-out folds cost nothing measurable and can substantially reduce computation time
(Appendix~\ref{app:context}), and fewer feature subsets trade marginal accuracy
for a significant speedup (Appendix~\ref{app:sens}). Every speedup of the backbone's
forward pass, such as the reduced KV cache of the TabPFN-3 report
\citep{grinsztajn2026tabpfn3}, lowers our cost directly.

\paragraph{Scale.}
ADBench's standard evaluation subsamples datasets above $10{,}000$
samples (Sec.~\ref{sec:setup}), so the largest reference set our
methods have seen holds $10{,}000$ samples. Every score is a set of
TabPFN forward passes over the whole reference set, so the cost grows
with its size, and larger sets are untested. Yet, TabPFN-3 handles
up to one million samples \citep{grinsztajn2026tabpfn3}, and future models will likely accommodate even more. Learned
contexts \citep{feuer2024tunetables} and test-time adaptations
\citep{ye2025closerlook}, are emerging options to extend the same methods to larger
tables. Three datasets in ADBench have more than 200 features. In these cases we used a fixed subset of 200 features
(Appendix~\ref{app:scope}). A more sophisticated treatment of wide tables can further improve our method.

\vspace{-0pt}
\section{Conclusion}
\vspace{-0pt}

\label{sec:conclusion}

In this paper we show that pretrained PFNs, once adapted
to the task, make a strong anomaly detector. Our
approach first embeds the data with the frozen model on random feature
subsets and softly cleans the reference set. Under contamination it
already reaches a higher mean AUROC than every classical and deep
baseline in our comparison. FOCUS, our trust-weighted compactness
fine-tune, reaches the highest mean AUROC in both clean and contaminated regimes and is
significantly ahead of every baseline in the contaminated regime.

For the PFN community our results are evidence that in-context
representations carry well beyond their pretraining objective. We also
show that our method of extracting meaningful representations for
anomaly detection, applied without any tuning or change, significantly
improves four other tabular foundation models as well. None of the
examined models was pretrained for anomaly detection, yet all of them
exhibited strong anomaly-detection capabilities when adapted
appropriately. This suggests that such capabilities are broadly
accessible across in-context tabular models, even for tasks that were
never used for pretraining.

\clearpage
\subsection*{AI use statement}
In this work, we used generative AI tools (LLM-based coding agents
operating under continuous author direction) for tasks with required
disclosure: implementing the methods and experiment infrastructure
specified by the authors, assisting in the interpretation of results,
and assisting with translation. We have not used generative AI tools to
propose or refine hypotheses, to design or provide feedback on the
research methodology or experiments, to generate synthetic datasets, to
clean and reformat datasets, or to support qualitative or thematic data
analysis; the methods and the experimental design are due to the
authors. The remaining required-disclosure tasks, developing theoretical
models or conceptual frameworks, formulating mathematical claims,
providing ingredients for proving them, and writing proofs, are not
applicable to this work. Additionally, we used generative AI tools for tasks with
recommended disclosure: creating and editing software code, creating
scientific figures, editing the text of this paper following the
authors' outlines and directions, searching and summarizing related
literature, and formatting references. We have reviewed all AI-assisted
work: experiment code was validated against logged design decisions
before any of its results entered the paper, every reported number was
recomputed from stored per-run artifacts, all text was reviewed and
edited by the authors, and every citation was checked against its
source. We take responsibility for the final content of this work,
including text, claims, and artifacts produced with the aid of
generative AI.

\subsection*{Reproducibility statement}
Every experiment in this paper runs on five fixed seeds, and every
(method, dataset, seed) result is stored as an individual file; each
reported number is an aggregate of those files. Section~\ref{sec:setup}
summarizes and Appendix~\ref{app:protocol} specifies the datasets, the two
regimes, the preprocessing, and the evaluation protocol, including the
data-leak correction to the original generator. Appendix~\ref{app:config} lists the complete configuration: the backbone
version, the readout layers, the fine-tuning schedule, and every constant, and
Appendix~\ref{app:datasets} gives per-dataset tables for all methods. We release
the code needed to reproduce every result in this paper in a public
repository
(\url{https://github.com/Maximilianb1/zen-focus-tabular-anomaly-detection}).

\subsection*{Ethics statement}
This work uses only publicly available, anonymized benchmark datasets
and involves no human subjects or personally identifiable information.
The contribution is methodological. Anomaly detectors do affect people
once deployed, in fraud screening or medical triage among others, and
any such deployment should audit error rates across the populations it
touches; nothing in our method removes that obligation. We are aware of
no other ethical concerns.

\bibliography{references}
\bibliographystyle{iclr2027_conference}

\appendix
% The appendix allows three top floats and four floats per page.
\setcounter{topnumber}{3}
\setcounter{totalnumber}{4}

\section{Experimental protocol in full}
\label{app:protocol}

\subsection{Splits and subsampling}
\label{app:splits}
This appendix completes the protocol of Sec.~\ref{sec:setup}. In the
clean regime the normal-class indices are shuffled, and the first
$\lfloor 0.7\,n\rfloor$ of them form the training set, where $n$ is the
dataset's size after subsampling. The test set holds every remaining
normal sample and every anomaly. When a dataset has too few normal
samples to fill that count, the fraction steps down by $0.1$ until it
fits; this keeps the split arithmetic of the standard pipeline. Eight
datasets with anomaly rates above $30\%$ step down to $0.6$. On the
most contaminated of them the clean test set is left with few normal
samples; SpamBase ($39.9\%$ anomalies) keeps four. Every method uses
the same split, so no comparison is affected. In the contaminated
regime the split is stratified $70/30$ and preserves the anomaly rate
on both sides.

Datasets above $10{,}000$ samples are subsampled to exactly $10{,}000$
before splitting. The draw is without replacement, separately from the
normal and the anomaly pool: the anomaly count is
$\mathrm{round}(\text{natural rate}\times10{,}000)$, floored at two and
capped at the true anomaly count, and normal samples fill the rest.
This preserves the natural anomaly rate exactly rather than in
expectation, and it can never duplicate a sample.

\subsection{The resampling leak in the official ADBench pipeline}
\label{app:leak}
ADBench's official generator resamples every dataset below $1{,}000$
samples to exactly $1{,}000$ by drawing samples with replacement, before the
train/test split \citep{han2022adbench}. Copies of one physical sample can
therefore land on both sides of the split. Twelve of the 47 datasets
are affected. Table~\ref{tab:leak} reports, for each of them, the
fraction of test samples whose byte-identical twin sits in the training
set under that protocol, measured by replicating the official code path
on the real data over seeds 0--4. Natural duplicate samples in the raw
data are not counted; without the resampling step the fraction is
structurally zero. In the contaminated regime, on average $88.5\%$ of
all test samples have a training twin. In the clean regime the training
set is normal-only, so anomalies cannot leak; instead $91.5\%$ of the
test \emph{normal} samples have one. For a distance- or density-based
scorer a twin sits at distance zero, which drives the affected normal
test scores toward their minimum and widens the apparent
normal/anomaly separation. The inflation therefore acts upward on
exactly the detector families these benchmarks rank.

We evaluate the affected datasets at their natural size instead: no
resampling, no duplication. The $10{,}000$-sample subsample of
Sec.~\ref{sec:setup} is unrelated to this leak; it draws without
replacement and duplicates nothing. One consequence is deliberate: our
numbers are not comparable cell-for-cell with numbers produced under
the official pipeline, on these twelve datasets by construction and in
any aggregate through them.

\begin{table}[t]
\centering
% Source: final/reference_checks/adbench_oversampling_leak.md (F78).
\caption{The twelve datasets the official resampling affects, with the
fraction of test samples whose byte-identical twin lands in the
training set under that protocol (mean over seeds 0--4). Our protocol
runs these datasets at their natural size, where the fraction is zero.}
\label{tab:leak}
\medskip
\small
\begin{tabular}{lrrr}
\toprule
Dataset & $n$ & clean: \% test normals & contam.: \% test samples\\
\midrule
breastw & 683 & 68.2 & 59.9\\
glass & 214 & 96.0 & 95.1\\
Hepatitis & 80 & 100.0 & 100.0\\
Ionosphere & 351 & 93.2 & 87.2\\
Lymphography & 148 & 99.8 & 99.5\\
Pima & 768 & 72.0 & 58.1\\
Stamps & 340 & 91.4 & 88.7\\
vertebral & 240 & 96.8 & 94.6\\
WBC & 223 & 95.8 & 96.4\\
WDBC & 367 & 86.4 & 85.7\\
wine & 129 & 99.4 & 99.4\\
WPBC & 198 & 99.4 & 97.1\\
\midrule
\textit{Mean} & & 91.5 & 88.5\\
\bottomrule
\end{tabular}
\end{table}

\subsection{Notes on specific datasets and detectors}
\label{app:scope}
TabPFN-3 is characterized up to 200 input features
\citep{grinsztajn2026tabpfn3}, and past that the library switches to an
internal ensemble of estimators. Our fine-tuning implementation trains a
single shared module, so that ensemble would multiply the cost and break
the design. On the
three datasets with more than 200 features (census, InternetAds, speech),
ZEN and FOCUS therefore use a fixed subset of 200 features, drawn once with
constant features excluded; every other method uses every feature.
uTabPFN builds its density one feature at a time, so its cost grows with
the feature count. On the same three datasets a single seed takes about
five to seven GPU-hours, and we did not run all five. Its aggregates therefore
cover 44 of 47 datasets, marked wherever they
appear. Four PyOD detectors (COPOD, ECOD, SOD, and COF) compute part of
their statistics over the batch being scored, which is how the library
and ADBench run them; every other method, ours included, scores each
test sample from training-set state alone. CBLOF fails on a small
number of seeds when its clustering cannot separate the data; its
dataset means average the completed seeds.

\section{Datasets and per-dataset results}
\label{app:datasets}

\subsection{Dataset statistics}
\label{app:dstats}
Table~\ref{tab:datasets} lists the 47 datasets with their natural size
$n$, dimensionality $d$, and anomaly rate, computed from the benchmark
files our loader reads. The markers flag the three protocol facts that
touch individual datasets: the sub-$1{,}000$ sets the official
resampling would affect (Appendix~\ref{app:leak}), the sets subsampled
to $10{,}000$ samples, and the three sets where ZEN and FOCUS use a fixed
200-feature subset (Appendix~\ref{app:scope}).

\begin{table}[t]
\centering
% Body: final/figures/paper_tab_datasets.tex (build_appendix_v1.py).
\caption{The 47 ADBench classical datasets: natural size $n$,
dimensionality $d$, and anomaly rate. $^{\dagger}$\,below $1{,}000$
samples (affected by the official pipeline's resampling; run at natural
size here). $^{\ddagger}$\,subsampled to $10{,}000$ samples.
$^{\S}$\,ZEN and FOCUS use a fixed 200-feature subset.}
\label{tab:datasets}
\medskip
\small
\begin{tabular}{lrrr@{\hspace{18pt}}lrrr}
\toprule
Dataset & $n$ & $d$ & anom.\,\% & Dataset & $n$ & $d$ & anom.\,\%\\
\midrule
ALOI$^{\ddagger}$ & 49,534 & 27 & 3.0 & musk & 3,062 & 166 & 3.2\\
annthyroid & 7,200 & 6 & 7.4 & optdigits & 5,216 & 64 & 2.9\\
backdoor$^{\ddagger}$ & 95,329 & 196 & 2.4 & PageBlocks & 5,393 & 10 & 9.5\\
breastw$^{\dagger}$ & 683 & 9 & 35.0 & pendigits & 6,870 & 16 & 2.3\\
campaign$^{\ddagger}$ & 41,188 & 62 & 11.3 & Pima$^{\dagger}$ & 768 & 8 & 34.9\\
cardio & 1,831 & 21 & 9.6 & satellite & 6,435 & 36 & 31.6\\
Cardiotocography & 2,114 & 21 & 22.0 & satimage-2 & 5,803 & 36 & 1.2\\
celeba$^{\ddagger}$ & 202,599 & 39 & 2.2 & shuttle$^{\ddagger}$ & 49,097 & 9 & 7.2\\
census$^{\ddagger}$$^{\S}$ & 299,285 & 500 & 6.2 & skin$^{\ddagger}$ & 245,057 & 3 & 20.8\\
cover$^{\ddagger}$ & 286,048 & 10 & 0.96 & smtp$^{\ddagger}$ & 95,156 & 3 & 0.03\\
donors$^{\ddagger}$ & 619,326 & 10 & 5.9 & SpamBase & 4,207 & 57 & 39.9\\
fault & 1,941 & 27 & 34.7 & speech$^{\S}$ & 3,686 & 400 & 1.6\\
fraud$^{\ddagger}$ & 284,807 & 29 & 0.17 & Stamps$^{\dagger}$ & 340 & 9 & 9.1\\
glass$^{\dagger}$ & 214 & 7 & 4.2 & thyroid & 3,772 & 6 & 2.5\\
Hepatitis$^{\dagger}$ & 80 & 19 & 16.2 & vertebral$^{\dagger}$ & 240 & 6 & 12.5\\
http$^{\ddagger}$ & 567,498 & 3 & 0.39 & vowels & 1,456 & 12 & 3.4\\
InternetAds$^{\S}$ & 1,966 & 1555 & 18.7 & Waveform & 3,443 & 21 & 2.9\\
Ionosphere$^{\dagger}$ & 351 & 32 & 35.9 & WBC$^{\dagger}$ & 223 & 9 & 4.5\\
landsat & 6,435 & 36 & 20.7 & WDBC$^{\dagger}$ & 367 & 30 & 2.7\\
letter & 1,600 & 32 & 6.2 & Wilt & 4,819 & 5 & 5.3\\
Lymphography$^{\dagger}$ & 148 & 18 & 4.1 & wine$^{\dagger}$ & 129 & 13 & 7.8\\
magic.gamma$^{\ddagger}$ & 19,020 & 10 & 35.2 & WPBC$^{\dagger}$ & 198 & 33 & 23.7\\
mammography$^{\ddagger}$ & 11,183 & 6 & 2.3 & yeast & 1,484 & 8 & 34.2\\
mnist & 7,603 & 100 & 9.2 &  & & & \\
\bottomrule
\end{tabular}
\end{table}

\subsection{Per-dataset AUROC}
\label{app:perdataset}
Tables~\ref{tab:pdnoisy1} through~\ref{tab:pdclean2} give every
compared method's per-dataset AUROC, averaged over the five seeds. The
bottom rows reproduce the means of Table~\ref{tab:main}.

\begin{table}[p]
\centering
\caption{Per-dataset AUROC ($\times100$, five-seed means),
contaminated regime. Methods ordered by mean AUROC; ours marked,
markers as in Table~\ref{tab:main}. The 19 methods are split over two
tables for width: this table holds the stronger half,
Table~\ref{tab:pdnoisy2} the rest, same 47 datasets.}
% Body: anon-release paper_assets/table_per_dataset_{noisy,clean}_{1,2}.tex (build_paper_assets.py), this and the next three tables.
\label{tab:pdnoisy1}
\medskip
\scriptsize
\setlength{\tabcolsep}{3pt}
\begin{tabular}{lrrrrrrrrrr}
\toprule
Dataset & FOCUS (ours) & ZEN (ours) & IForest & TabPFN-PL$^{\mathrm{t}}$ & COPOD & HBOS & DTE$^{\mathrm{d}}$ & raw $k$NN & ECOD & CBLOF\\
\midrule
ALOI & 56.0 & 55.5 & 53.9 & 61.5 & 52.3 & 54.6 & 55.8 & 59.0 & 54.7 & 56.1\\
annthyroid & 85.9 & 84.3 & 82.9 & 68.7 & 78.0 & 62.9 & 65.1 & 72.3 & 79.1 & 65.6\\
backdoor & 89.1 & 87.6 & 72.9 & 91.7 & 78.5 & 70.3 & 90.2 & 76.6 & 84.2 & 77.4\\
breastw & 98.1 & 97.7 & 98.9 & 97.7 & 99.4 & 98.6 & 89.9 & 98.4 & 99.1 & 97.1\\
campaign & 77.4 & 78.6 & 71.2 & 80.6 & 78.7 & 79.3 & 72.2 & 69.1 & 77.5 & 65.9\\
cardio & 90.0 & 86.9 & 93.1 & 95.0 & 91.9 & 84.6 & 80.3 & 80.8 & 94.2 & 85.4\\
Cardiotocography & 62.6 & 62.8 & 68.8 & 65.6 & 65.7 & 59.8 & 60.1 & 55.0 & 78.5 & 62.9\\
celeba & 76.2 & 71.7 & 69.6 & 73.9 & 74.7 & 75.1 & 68.2 & 62.4 & 75.2 & 61.1\\
census & 63.2 & 62.4 & 59.9 & 60.9 & 66.6 & 62.8 & 65.5 & 64.1 & 65.7 & 58.5\\
cover & 89.4 & 91.8 & 89.6 & 93.3 & 90.2 & 78.5 & 93.6 & 90.2 & 93.4 & 90.0\\
donors & 87.5 & 86.8 & 77.6 & 68.6 & 82.5 & 78.7 & 74.3 & 81.8 & 89.4 & 74.0\\
fault & 71.0 & 70.1 & 56.7 & 54.4 & 46.6 & 56.6 & 60.9 & 73.8 & 47.8 & 65.2\\
fraud & 95.6 & 95.9 & 94.4 & 93.4 & 93.7 & 95.0 & 94.0 & 95.8 & 94.0 & 93.0\\
glass & 84.3 & 85.4 & 80.2 & 75.3 & 72.9 & 81.4 & 70.5 & 87.0 & 68.8 & 74.5\\
Hepatitis & 75.0 & 76.2 & 76.5 & 84.2 & 82.2 & 77.5 & 75.2 & 74.8 & 77.2 & 65.8\\
http & 99.3 & 99.4 & 100.0 & 91.5 & 99.0 & 99.5 & 99.5 & 1.6 & 98.0 & 99.5\\
InternetAds & 67.6 & 67.7 & 68.0 & 70.3 & 67.3 & 68.7 & 61.8 & 71.9 & 67.4 & 69.7\\
Ionosphere & 89.5 & 89.6 & 81.9 & 88.4 & 77.2 & 61.5 & 95.3 & 90.0 & 70.8 & 86.4\\
landsat & 55.5 & 58.0 & 49.3 & 47.3 & 41.2 & 56.5 & 47.8 & 59.3 & 35.7 & 55.8\\
letter & 87.4 & 87.6 & 63.4 & 64.1 & 55.5 & 60.2 & 74.8 & 86.8 & 56.3 & 73.8\\
Lymphography & 99.5 & 99.5 & 100.0 & 96.5 & 100.0 & 99.8 & 99.5 & 100.0 & 99.5 & 100.0\\
magic.gamma & 75.6 & 77.9 & 72.8 & 65.5 & 68.0 & 70.4 & 76.0 & 83.0 & 63.7 & 76.0\\
mammography & 84.8 & 84.5 & 85.8 & 82.2 & 90.3 & 84.8 & 82.2 & 84.6 & 90.4 & 81.8\\
mnist & 86.3 & 84.6 & 78.8 & 87.5 & 77.7 & 61.8 & 77.2 & 82.8 & 75.1 & 79.7\\
musk & 100.0 & 100.0 & 99.8 & 99.4 & 94.1 & 100.0 & 84.0 & 84.7 & 95.0 & 100.0\\
optdigits & 89.5 & 92.7 & 75.4 & 36.5 & 67.6 & 82.1 & 54.4 & 39.9 & 59.5 & 83.4\\
PageBlocks & 96.5 & 94.9 & 90.1 & 89.1 & 87.8 & 80.5 & 90.4 & 85.2 & 91.8 & 87.1\\
pendigits & 88.4 & 88.4 & 93.2 & 92.5 & 90.7 & 92.8 & 79.9 & 74.6 & 92.7 & 96.5\\
Pima & 68.8 & 69.7 & 70.0 & 64.6 & 64.8 & 68.3 & 63.6 & 73.6 & 60.3 & 67.6\\
satellite & 71.4 & 67.9 & 70.0 & 68.8 & 62.5 & 75.4 & 68.3 & 67.3 & 57.4 & 71.0\\
satimage-2 & 99.8 & 99.7 & 99.4 & 95.5 & 97.3 & 97.8 & 99.2 & 96.7 & 96.1 & 99.9\\
shuttle & 80.3 & 72.5 & 99.7 & 99.4 & 99.6 & 99.0 & 98.0 & 72.4 & 99.3 & 74.0\\
skin & 79.5 & 74.2 & 67.2 & 40.7 & 46.6 & 58.8 & 80.5 & 71.0 & 48.4 & 63.6\\
smtp & 84.8 & 83.7 & 87.3 & 72.0 & 87.7 & 82.5 & 83.1 & 89.9 & 82.5 & 82.4\\
SpamBase & 69.7 & 76.6 & 62.6 & 40.8 & 69.2 & 65.0 & 54.6 & 54.8 & 65.8 & 56.5\\
speech & 48.4 & 47.6 & 49.5 & 48.3 & 49.3 & 46.8 & 51.9 & 48.9 & 46.8 & 46.6\\
Stamps & 88.2 & 88.5 & 90.0 & 87.9 & 93.2 & 90.6 & 78.0 & 89.3 & 86.7 & 72.6\\
thyroid & 98.3 & 98.0 & 98.0 & 95.7 & 94.1 & 95.5 & 95.0 & 94.9 & 97.9 & 92.8\\
vertebral & 32.7 & 32.0 & 37.8 & 41.9 & 36.4 & 30.9 & 42.7 & 35.0 & 45.3 & 39.3\\
vowels & 96.0 & 97.9 & 70.0 & 74.1 & 42.6 & 64.6 & 83.3 & 97.0 & 51.0 & 89.9\\
Waveform & 79.2 & 79.0 & 65.9 & 73.2 & 70.8 & 65.4 & 61.7 & 74.3 & 56.5 & 69.7\\
WBC & 99.6 & 99.6 & 99.5 & 99.3 & 99.4 & 99.2 & 97.2 & 99.4 & 99.4 & 98.6\\
WDBC & 96.0 & 95.9 & 98.1 & 97.4 & 99.1 & 99.2 & 93.6 & 97.0 & 96.5 & 97.7\\
Wilt & 69.1 & 64.7 & 46.4 & 44.1 & 34.2 & 38.3 & 31.7 & 45.7 & 39.9 & 33.0\\
wine & 85.0 & 85.2 & 86.5 & 94.4 & 92.4 & 90.7 & 78.1 & 88.5 & 81.5 & 30.7\\
WPBC & 49.3 & 48.4 & 49.9 & 51.6 & 53.4 & 56.4 & 51.8 & 55.3 & 44.3 & 53.6\\
yeast & 43.5 & 45.1 & 38.2 & 41.3 & 36.2 & 40.1 & 40.9 & 37.6 & 43.4 & 43.8\\
\midrule
\textit{Mean} & 80.0 & 79.7 & 76.4 & 74.6 & 74.4 & 74.4 & 74.3 & 73.9 & 73.9 & 73.7\\
\bottomrule
\end{tabular}
\end{table}

\begin{table}[p]
\centering
\caption{Per-dataset AUROC ($\times100$, five-seed means),
contaminated regime: the remaining methods, continuing
Table~\ref{tab:pdnoisy1} over the same 47 datasets.
$^{\dagger}$uTabPFN covers 44 of 47 datasets.}
\label{tab:pdnoisy2}
\medskip
\scriptsize
\setlength{\tabcolsep}{3pt}
\begin{tabular}{lrrrrrrrrr}
\toprule
Dataset & PCA & TCCM$^{\mathrm{d}}$ & DeepSVDD$^{\mathrm{d}}$ & OCSVM & SOD & uTabPFN$^{\mathrm{t}}$$^{\dagger}$ & COF & LODA & LOF\\
\midrule
ALOI & 56.5 & 57.3 & 56.2 & 55.1 & 61.0 & 61.0 & 66.1 & 50.4 & 66.6\\
annthyroid & 66.1 & 74.3 & 69.3 & 59.1 & 79.3 & 87.2 & 70.7 & 44.4 & 71.0\\
backdoor & 81.0 & 90.7 & 78.6 & 83.3 & 69.8 & 41.1 & 72.1 & 76.3 & 76.9\\
breastw & 94.6 & 91.2 & 91.5 & 87.2 & 94.0 & 76.8 & 44.5 & 99.1 & 46.8\\
campaign & 74.6 & 73.0 & 70.4 & 67.8 & 70.3 & 50.3 & 56.3 & 57.2 & 57.3\\
cardio & 95.0 & 69.5 & 89.1 & 93.9 & 67.0 & 56.6 & 60.8 & 88.0 & 67.4\\
Cardiotocography & 74.8 & 78.9 & 73.1 & 78.1 & 51.8 & 52.0 & 55.3 & 75.0 & 57.4\\
celeba & 77.6 & 67.6 & 71.7 & 68.8 & 50.1 & 43.8 & 43.1 & 59.2 & 42.2\\
census & 65.7 & 59.0 & 65.3 & 53.4 & 60.1 & -- & 42.1 & 44.3 & 47.4\\
cover & 94.1 & 93.2 & 85.6 & 92.9 & 74.9 & 66.5 & 80.6 & 89.1 & 86.6\\
donors & 82.7 & 80.0 & 79.1 & 72.3 & 62.2 & 63.2 & 70.8 & 47.3 & 59.0\\
fault & 49.3 & 61.2 & 51.9 & 53.0 & 66.7 & 56.3 & 60.7 & 48.9 & 58.7\\
fraud & 94.8 & 94.2 & 93.1 & 94.1 & 95.2 & 95.9 & 93.5 & 93.9 & 95.7\\
glass & 69.8 & 76.9 & 60.2 & 36.9 & 84.4 & 78.1 & 85.3 & 71.8 & 80.9\\
Hepatitis & 78.0 & 82.5 & 77.8 & 71.5 & 62.7 & 74.5 & 70.0 & 61.3 & 74.5\\
http & 99.7 & 84.8 & 99.8 & 99.5 & 39.0 & 97.0 & 92.7 & 6.6 & 26.8\\
InternetAds & 60.4 & 68.7 & 59.2 & 68.9 & 61.8 & -- & 64.6 & 58.6 & 64.0\\
Ionosphere & 77.0 & 86.3 & 72.8 & 73.7 & 85.1 & 74.2 & 85.2 & 77.3 & 87.6\\
landsat & 35.1 & 52.8 & 38.4 & 34.9 & 60.9 & 56.4 & 53.5 & 37.0 & 55.0\\
letter & 51.2 & 82.1 & 50.3 & 48.1 & 86.3 & 30.9 & 80.9 & 52.0 & 85.4\\
Lymphography & 99.5 & 99.8 & 99.1 & 99.8 & 95.6 & 5.8 & 97.7 & 68.8 & 99.5\\
magic.gamma & 66.9 & 70.8 & 63.5 & 60.1 & 76.7 & 84.1 & 66.5 & 68.2 & 68.4\\
mammography & 89.9 & 82.3 & 81.6 & 86.6 & 79.6 & 85.0 & 76.7 & 83.7 & 76.3\\
mnist & 85.2 & 74.0 & 70.8 & 83.0 & 58.3 & 61.0 & 70.6 & 74.2 & 68.8\\
musk & 100.0 & 70.8 & 96.0 & 92.4 & 76.5 & 96.2 & 44.5 & 95.0 & 35.8\\
optdigits & 51.3 & 60.6 & 51.8 & 55.0 & 57.1 & 84.0 & 51.3 & 55.7 & 56.3\\
PageBlocks & 91.3 & 79.5 & 90.2 & 90.1 & 78.2 & 77.5 & 72.5 & 83.4 & 72.9\\
pendigits & 93.6 & 75.5 & 85.5 & 94.9 & 65.1 & 79.1 & 44.7 & 91.4 & 42.6\\
Pima & 65.1 & 68.7 & 64.0 & 61.8 & 57.4 & 57.2 & 65.0 & 63.7 & 63.6\\
satellite & 58.9 & 63.5 & 60.8 & 58.9 & 64.7 & 69.7 & 54.9 & 60.5 & 54.5\\
satimage-2 & 97.5 & 87.9 & 97.2 & 98.1 & 86.7 & 95.8 & 57.7 & 98.2 & 54.9\\
shuttle & 99.0 & 58.2 & 98.7 & 98.2 & 69.2 & 97.5 & 52.4 & 81.6 & 52.8\\
skin & 12.2 & 65.2 & 47.8 & 48.5 & 59.0 & 69.8 & 41.7 & 42.9 & 44.9\\
smtp & 72.6 & 69.3 & 72.7 & 69.3 & 58.5 & 82.3 & 74.2 & 68.1 & 93.7\\
SpamBase & 54.9 & 47.7 & 54.0 & 52.3 & 52.3 & 61.5 & 41.4 & 43.8 & 44.5\\
speech & 46.7 & 50.0 & 54.0 & 46.0 & 50.2 & -- & 45.8 & 47.8 & 49.0\\
Stamps & 90.6 & 72.1 & 90.8 & 82.9 & 80.9 & 74.0 & 81.0 & 88.4 & 69.6\\
thyroid & 95.8 & 94.0 & 96.4 & 86.8 & 91.6 & 96.7 & 87.3 & 78.0 & 85.7\\
vertebral & 44.7 & 44.4 & 37.8 & 42.9 & 40.1 & 35.8 & 41.0 & 30.7 & 37.0\\
vowels & 51.2 & 91.0 & 53.2 & 48.1 & 90.8 & 90.6 & 95.9 & 64.2 & 93.4\\
Waveform & 60.1 & 51.6 & 54.4 & 51.5 & 63.4 & 64.8 & 69.6 & 56.4 & 71.8\\
WBC & 98.8 & 98.6 & 98.1 & 99.5 & 97.0 & 44.6 & 99.5 & 99.0 & 92.1\\
WDBC & 98.0 & 97.7 & 96.2 & 97.5 & 94.6 & 98.1 & 96.3 & 97.4 & 97.2\\
Wilt & 37.7 & 31.0 & 34.6 & 36.1 & 60.6 & 69.7 & 55.0 & 34.8 & 52.4\\
wine & 85.2 & 73.3 & 65.7 & 78.1 & 65.9 & 66.7 & 90.0 & 88.9 & 91.5\\
WPBC & 46.5 & 45.6 & 45.7 & 45.4 & 51.5 & 58.1 & 53.4 & 51.6 & 52.9\\
yeast & 38.1 & 47.7 & 39.7 & 40.7 & 45.2 & 37.5 & 43.8 & 46.3 & 44.3\\
\midrule
\textit{Mean} & 72.5 & 72.2 & 70.9 & 70.1 & 69.1 & 68.3 & 66.4 & 66.0 & 65.4\\
\bottomrule
\end{tabular}
\end{table}

\begin{table}[p]
\centering
\caption{Per-dataset AUROC ($\times100$, five-seed means), clean
regime. Methods ordered by mean AUROC; ours marked, markers as in
Table~\ref{tab:main}. The 19 methods are split over two tables for
width: this table holds the stronger half, Table~\ref{tab:pdclean2}
the rest, same 47 datasets.}
\label{tab:pdclean1}
\medskip
\scriptsize
\setlength{\tabcolsep}{3pt}
\begin{tabular}{lrrrrrrrrrr}
\toprule
Dataset & FOCUS (ours) & ZEN (ours) & raw $k$NN & LOF & CBLOF & DTE$^{\mathrm{d}}$ & TCCM$^{\mathrm{d}}$ & IForest & TabPFN-PL$^{\mathrm{t}}$ & HBOS\\
\midrule
ALOI & 67.7 & 60.0 & 59.2 & 66.8 & 55.2 & 55.1 & 56.4 & 54.5 & 61.4 & 53.6\\
annthyroid & 91.5 & 90.2 & 86.2 & 91.5 & 73.8 & 73.4 & 83.6 & 91.2 & 82.6 & 70.8\\
backdoor & 98.2 & 97.4 & 94.5 & 93.4 & 83.1 & 92.4 & 92.9 & 76.7 & 86.7 & 65.7\\
breastw & 99.1 & 99.6 & 99.6 & 79.0 & 99.5 & 98.5 & 98.3 & 99.8 & 99.3 & 99.6\\
campaign & 77.9 & 79.1 & 70.0 & 64.8 & 68.9 & 74.9 & 75.1 & 73.5 & 82.5 & 80.0\\
cardio & 93.2 & 91.3 & 96.3 & 94.9 & 95.9 & 96.4 & 95.4 & 95.1 & 94.0 & 87.3\\
Cardiotocography & 74.8 & 71.4 & 78.8 & 79.9 & 79.0 & 78.4 & 82.6 & 79.9 & 80.4 & 70.7\\
celeba & 67.1 & 67.6 & 66.7 & 45.6 & 69.2 & 73.5 & 70.5 & 71.4 & 77.5 & 77.7\\
census & 65.2 & 64.6 & 66.6 & 49.4 & 62.2 & 68.2 & 60.4 & 59.1 & 65.7 & 65.0\\
cover & 98.7 & 98.1 & 96.8 & 98.8 & 94.0 & 97.9 & 92.8 & 85.1 & 90.8 & 77.4\\
donors & 99.7 & 99.2 & 100.0 & 95.6 & 93.1 & 96.4 & 91.9 & 90.2 & 90.4 & 82.8\\
fault & 79.6 & 75.5 & 80.0 & 64.3 & 71.2 & 70.6 & 70.4 & 65.0 & 67.2 & 64.0\\
fraud & 92.3 & 92.7 & 92.1 & 93.9 & 89.7 & 89.9 & 90.2 & 89.3 & 88.8 & 89.5\\
glass & 88.9 & 85.3 & 88.7 & 73.0 & 89.0 & 74.3 & 69.9 & 83.1 & 85.7 & 81.9\\
Hepatitis & 79.6 & 81.8 & 82.0 & 80.4 & 71.6 & 77.2 & 86.9 & 79.6 & 85.5 & 82.5\\
http & 99.9 & 99.8 & 99.9 & 99.9 & 99.5 & 99.5 & 100.0 & 99.2 & 91.6 & 99.3\\
InternetAds & 70.3 & 66.7 & 78.3 & 86.0 & 72.6 & 80.0 & 70.7 & 39.2 & 83.2 & 54.5\\
Ionosphere & 96.6 & 96.2 & 97.7 & 94.7 & 98.6 & 97.6 & 97.6 & 92.0 & 82.0 & 77.8\\
landsat & 76.5 & 76.2 & 76.7 & 76.2 & 69.2 & 51.1 & 59.9 & 58.5 & 57.3 & 68.2\\
letter & 96.7 & 95.1 & 89.2 & 88.7 & 78.2 & 81.2 & 87.5 & 63.3 & 65.1 & 59.7\\
Lymphography & 99.1 & 99.0 & 99.1 & 98.6 & 99.1 & 99.0 & 99.1 & 99.3 & 95.4 & 99.5\\
magic.gamma & 85.8 & 86.6 & 85.3 & 81.1 & 80.9 & 81.3 & 77.0 & 77.5 & 72.0 & 73.4\\
mammography & 87.8 & 87.6 & 88.4 & 85.9 & 86.4 & 88.7 & 88.9 & 87.9 & 84.9 & 85.9\\
mnist & 93.5 & 94.7 & 93.5 & 95.5 & 87.9 & 89.5 & 87.6 & 85.9 & 91.2 & 66.7\\
musk & 100.0 & 100.0 & 100.0 & 100.0 & 100.0 & 100.0 & 100.0 & 91.7 & 84.4 & 100.0\\
optdigits & 99.2 & 98.2 & 99.5 & 99.7 & 90.9 & 85.2 & 88.2 & 83.9 & 64.2 & 87.5\\
PageBlocks & 98.2 & 97.6 & 95.3 & 96.2 & 92.1 & 95.6 & 93.9 & 92.5 & 89.8 & 80.9\\
pendigits & 99.4 & 99.4 & 99.9 & 99.4 & 97.8 & 97.8 & 98.8 & 96.5 & 90.8 & 93.8\\
Pima & 72.7 & 72.4 & 75.1 & 71.7 & 72.2 & 69.3 & 61.0 & 72.3 & 71.1 & 72.1\\
satellite & 84.7 & 85.8 & 87.7 & 85.0 & 85.0 & 80.0 & 82.5 & 81.3 & 73.8 & 86.6\\
satimage-2 & 99.8 & 99.9 & 99.9 & 99.7 & 99.8 & 99.9 & 99.9 & 99.5 & 92.7 & 98.2\\
shuttle & 100.0 & 100.0 & 99.7 & 99.7 & 98.9 & 99.1 & 99.7 & 99.6 & 91.1 & 99.0\\
skin & 97.7 & 96.3 & 98.7 & 77.9 & 92.1 & 92.3 & 55.2 & 88.6 & 51.7 & 77.4\\
smtp & 74.3 & 76.2 & 88.5 & 85.5 & 82.9 & 80.7 & 69.9 & 85.8 & 71.4 & 84.9\\
SpamBase & 83.0 & 78.4 & 77.0 & 82.2 & 73.0 & 73.7 & 83.3 & 86.4 & 88.7 & 77.0\\
speech & 50.6 & 48.6 & 52.0 & 51.1 & 47.6 & 52.6 & 51.6 & 47.9 & 49.5 & 47.7\\
Stamps & 96.1 & 93.8 & 93.8 & 92.3 & 93.6 & 92.5 & 88.4 & 94.4 & 91.5 & 92.7\\
thyroid & 97.6 & 98.0 & 97.9 & 98.3 & 96.6 & 97.8 & 97.2 & 98.8 & 94.6 & 98.4\\
vertebral & 47.0 & 45.3 & 43.7 & 46.3 & 50.9 & 46.1 & 50.3 & 42.0 & 52.3 & 38.1\\
vowels & 98.8 & 97.1 & 98.1 & 97.1 & 93.8 & 93.5 & 94.4 & 78.8 & 76.9 & 70.2\\
Waveform & 73.6 & 78.0 & 77.7 & 77.4 & 74.3 & 67.7 & 64.4 & 71.9 & 80.9 & 70.0\\
WBC & 99.2 & 99.2 & 99.4 & 96.9 & 99.4 & 97.5 & 99.3 & 99.4 & 98.0 & 99.0\\
WDBC & 99.7 & 99.6 & 99.5 & 99.8 & 99.4 & 98.8 & 99.5 & 99.5 & 96.2 & 99.0\\
Wilt & 90.5 & 91.7 & 54.0 & 61.4 & 37.1 & 32.9 & 37.5 & 49.9 & 54.6 & 43.3\\
wine & 94.4 & 93.7 & 96.7 & 96.3 & 96.3 & 96.1 & 95.5 & 92.3 & 83.1 & 90.4\\
WPBC & 53.3 & 53.8 & 57.3 & 56.8 & 52.2 & 54.9 & 48.6 & 53.7 & 52.9 & 57.9\\
yeast & 46.9 & 46.2 & 42.9 & 46.3 & 45.6 & 42.9 & 52.5 & 39.8 & 44.4 & 40.9\\
\midrule
\textit{Mean} & 85.9 & 85.2 & 85.1 & 82.9 & 81.7 & 81.5 & 80.8 & 79.6 & 78.8 & 77.4\\
\bottomrule
\end{tabular}
\end{table}

\begin{table}[p]
\centering
\caption{Per-dataset AUROC ($\times100$, five-seed means), clean
regime: the remaining methods, continuing Table~\ref{tab:pdclean1}
over the same 47 datasets. $^{\dagger}$uTabPFN covers 44 of 47
datasets.}
\label{tab:pdclean2}
\medskip
\scriptsize
\setlength{\tabcolsep}{3pt}
\begin{tabular}{lrrrrrrrrr}
\toprule
Dataset & PCA & OCSVM & DeepSVDD$^{\mathrm{d}}$ & COPOD & ECOD & uTabPFN$^{\mathrm{t}}$$^{\dagger}$ & LODA & SOD & COF\\
\midrule
ALOI & 55.6 & 54.8 & 55.4 & 51.7 & 53.3 & 63.3 & 51.5 & 59.0 & 63.2\\
annthyroid & 80.9 & 66.1 & 84.9 & 77.4 & 78.7 & 93.7 & 58.7 & 71.9 & 64.1\\
backdoor & 69.4 & 84.7 & 58.7 & 78.4 & 84.2 & 36.5 & 66.9 & 69.5 & 73.7\\
breastw & 99.1 & 99.6 & 99.0 & 99.7 & 99.5 & 6.6 & 99.5 & 86.9 & 56.8\\
campaign & 77.0 & 69.1 & 72.9 & 78.1 & 77.0 & 53.1 & 55.7 & 65.5 & 52.5\\
cardio & 96.4 & 97.4 & 93.6 & 92.0 & 93.6 & 87.5 & 91.3 & 49.0 & 53.4\\
Cardiotocography & 82.8 & 85.2 & 79.5 & 66.2 & 78.5 & 82.4 & 80.1 & 35.5 & 41.6\\
celeba & 80.6 & 72.1 & 74.8 & 76.0 & 76.6 & 44.1 & 62.5 & 48.3 & 39.5\\
census & 71.0 & 56.1 & 69.5 & 66.5 & 65.6 & -- & 44.3 & 58.2 & 41.0\\
cover & 94.3 & 95.0 & 86.3 & 88.4 & 92.0 & 92.4 & 92.3 & 57.9 & 50.7\\
donors & 88.0 & 88.3 & 84.1 & 81.4 & 89.1 & 68.0 & 46.8 & 56.1 & 69.6\\
fault & 55.0 & 55.9 & 56.7 & 45.4 & 47.7 & 71.4 & 51.6 & 50.7 & 50.9\\
fraud & 90.1 & 89.9 & 88.6 & 88.9 & 88.9 & 92.4 & 89.7 & 91.8 & 91.5\\
glass & 73.9 & 47.4 & 64.5 & 78.1 & 72.2 & 88.7 & 71.4 & 76.5 & 82.2\\
Hepatitis & 84.8 & 81.5 & 76.5 & 82.4 & 81.7 & 76.5 & 63.6 & 46.6 & 46.6\\
http & 99.9 & 99.5 & 99.8 & 98.9 & 97.8 & 99.9 & 34.6 & 47.3 & 2.1\\
InternetAds & 80.1 & 71.3 & 77.7 & 68.3 & 68.3 & -- & 55.6 & 35.6 & 46.5\\
Ionosphere & 89.9 & 92.9 & 84.5 & 80.0 & 74.1 & 95.7 & 88.6 & 73.2 & 75.4\\
landsat & 39.4 & 37.2 & 41.6 & 41.1 & 35.8 & 75.2 & 41.0 & 45.3 & 50.7\\
letter & 53.6 & 51.8 & 50.0 & 56.1 & 57.2 & 18.4 & 54.3 & 78.5 & 81.7\\
Lymphography & 99.1 & 99.0 & 98.1 & 99.5 & 99.3 & 9.4 & 37.2 & 97.2 & 98.8\\
magic.gamma & 70.6 & 71.9 & 67.6 & 68.6 & 64.6 & 88.1 & 73.0 & 66.9 & 59.7\\
mammography & 90.5 & 89.9 & 85.4 & 90.5 & 90.6 & 87.8 & 89.5 & 78.2 & 67.8\\
mnist & 90.5 & 88.6 & 80.6 & 77.4 & 74.8 & 78.5 & 82.0 & 61.6 & 57.7\\
musk & 100.0 & 100.0 & 98.5 & 94.8 & 95.8 & 97.7 & 99.9 & 54.4 & 52.7\\
optdigits & 58.0 & 64.5 & 54.7 & 68.4 & 60.6 & 32.1 & 66.7 & 44.4 & 56.9\\
PageBlocks & 93.2 & 92.0 & 92.8 & 87.5 & 91.5 & 88.4 & 89.0 & 56.5 & 62.3\\
pendigits & 94.2 & 96.4 & 86.9 & 90.5 & 92.7 & 99.7 & 93.4 & 51.3 & 51.5\\
Pima & 69.9 & 67.7 & 67.2 & 64.7 & 60.1 & 63.9 & 67.6 & 50.1 & 54.9\\
satellite & 67.0 & 65.6 & 68.0 & 63.2 & 58.4 & 86.9 & 68.7 & 47.5 & 51.5\\
satimage-2 & 97.8 & 98.2 & 97.2 & 97.5 & 96.5 & 99.7 & 98.4 & 74.3 & 55.9\\
shuttle & 99.3 & 98.6 & 99.2 & 99.5 & 99.3 & 12.8 & 92.7 & 42.6 & 55.6\\
skin & 26.1 & 75.3 & 60.7 & 45.9 & 48.2 & 99.2 & 74.5 & 42.4 & 40.7\\
smtp & 70.5 & 66.2 & 68.1 & 87.4 & 81.1 & 83.8 & 66.5 & 60.6 & 77.9\\
SpamBase & 80.4 & 72.4 & 79.9 & 67.7 & 64.2 & 78.2 & 59.8 & 32.1 & 18.9\\
speech & 47.5 & 47.0 & 52.1 & 49.7 & 47.5 & -- & 48.1 & 43.9 & 37.3\\
Stamps & 93.9 & 91.3 & 94.3 & 93.2 & 88.2 & 94.9 & 92.3 & 57.2 & 36.8\\
thyroid & 98.0 & 93.6 & 98.5 & 93.7 & 97.6 & 98.5 & 91.4 & 87.3 & 51.8\\
vertebral & 44.1 & 42.7 & 42.1 & 35.6 & 42.3 & 48.0 & 36.3 & 38.0 & 29.9\\
vowels & 61.7 & 62.3 & 57.3 & 49.7 & 59.3 & 99.9 & 71.8 & 73.8 & 79.6\\
Waveform & 65.6 & 56.4 & 58.8 & 73.3 & 60.4 & 78.4 & 61.7 & 56.4 & 61.0\\
WBC & 99.2 & 99.4 & 98.4 & 99.3 & 99.3 & 7.2 & 97.9 & 91.1 & 91.9\\
WDBC & 99.5 & 99.7 & 98.9 & 99.5 & 97.1 & 99.7 & 99.1 & 87.2 & 92.8\\
Wilt & 39.2 & 35.8 & 35.3 & 34.7 & 39.7 & 90.7 & 33.7 & 47.6 & 47.7\\
wine & 92.3 & 93.7 & 83.1 & 85.6 & 73.2 & 97.2 & 87.7 & 30.8 & 16.3\\
WPBC & 51.9 & 51.8 & 46.4 & 53.2 & 50.9 & 53.9 & 54.8 & 48.2 & 44.4\\
yeast & 41.7 & 39.5 & 41.5 & 36.4 & 42.8 & 46.1 & 46.5 & 36.2 & 33.0\\
\midrule
\textit{Mean} & 76.7 & 75.6 & 74.7 & 74.5 & 74.2 & 72.0 & 69.8 & 58.7 & 55.7\\
\bottomrule
\end{tabular}
\end{table}

\subsection{AUPRC}
\label{app:auprc}
Table~\ref{tab:auprc} reports each method's mean AUPRC (average
precision, $\times100$) and its mean per-dataset AUPRC rank. AUPRC
depends strongly on the test anomaly rate, which spans $0.03\%$ to
$39.9\%$ across the contaminated regime's datasets
(Table~\ref{tab:datasets}) and runs higher still in the clean regime,
whose test sets hold every anomaly. A cross-dataset mean therefore
weights datasets very unevenly, and the rank column is the more
comparable summary. Every value is computed from stored score
vectors, and a method's AUPRC is reported only where those vectors
reproduce the method's official AUROC. FOCUS has the best mean AUPRC
in both regimes and the best mean rank under contamination. Its
contaminated lead over Isolation Forest is not significant
($p=0.08$), while its lead over raw-feature $k$NN is ($p=0.002$). In
the clean regime raw-feature $k$NN keeps the best AUPRC rank, and
FOCUS's lead over it is not significant ($p=0.52$).
FOCUS improves on ZEN's AUPRC in both regimes, significantly on clean
data ($p=0.03$) and not under contamination ($p=0.15$).

\begin{table}[t]
\centering
\caption{AUPRC ($\times100$) and mean per-dataset AUPRC rank over the
47 datasets, both regimes. Ranks cover the methods with values on all
47 datasets in that regime. A value appears only where the stored
score vectors reproduce that method's official AUROC (tolerance 0.25).
$^{\mathrm{a}}$\,TCCM's contaminated-regime score vectors were not
retained, so its AUPRC there is withheld (Appendix~\ref{app:baselines}).
$^{\dagger}$\,uTabPFN covers 44 datasets and is excluded from ranks.}
% Body: anon-release paper_assets/table_auprc.tex (build_paper_assets.py; F100 AUROC gate).
\label{tab:auprc}
\medskip
\small
\begin{tabular}{lrrrr}
\toprule
Method & \multicolumn{2}{c}{Contam.} & \multicolumn{2}{c}{Clean}\\
\cmidrule(lr){2-3}\cmidrule(lr){4-5}
 & AUPRC & rank & AUPRC & rank\\
\midrule
FOCUS (ours) & \textbf{41.7} & \textbf{5.6} & \textbf{78.3} & \underline{4.5}\\
ZEN (ours) & \underline{41.1} & \underline{6.3} & \underline{76.9} & 5.3\\
IForest & 39.1 & 7.7 & 62.3 & 9.1\\
PCA & 38.2 & 8.6 & 64.6 & 9.5\\
HBOS & 37.1 & 7.9 & 61.7 & 9.8\\
ECOD & 36.4 & 8.5 & 60.2 & 11.7\\
CBLOF & 35.8 & 8.4 & 68.1 & 7.6\\
COPOD & 35.7 & 8.7 & 58.3 & 12.0\\
DeepSVDD$^{\mathrm{d}}$ & 35.4 & 10.1 & 62.0 & 11.6\\
raw $k$NN & 35.3 & 7.9 & 76.2 & \textbf{4.3}\\
OCSVM & 34.5 & 10.1 & 63.8 & 10.1\\
DTE$^{\mathrm{d}}$ & 32.9 & 9.6 & 70.1 & 7.9\\
TabPFN-PL$^{\mathrm{t}}$ & 32.7 & 9.6 & 52.6 & 11.7\\
LODA & 30.3 & 10.9 & 56.3 & 12.6\\
uTabPFN$^{\mathrm{t}}$$^{\dagger}$ & 29.8 & -- & 63.4 & --\\
SOD & 28.9 & 10.6 & 44.2 & 15.4\\
COF & 26.9 & 11.4 & 43.0 & 15.4\\
LOF & 25.4 & 11.2 & 74.1 & 5.9\\
TCCM$^{\mathrm{d}}$$^{\mathrm{a}}$ & -- & -- & 70.9 & 6.6\\
\bottomrule
\end{tabular}
\end{table}

\subsection{Seed variability}
\label{app:seedstd}
Every number in Tables~\ref{tab:main} and \ref{tab:pdnoisy1}--\ref{tab:pdclean2}
is a mean over five seeds, each seed drawing a fresh split.
Table~\ref{tab:seedstd} reports the spread behind those means: the standard
deviation of each method's 47-dataset mean across the five seeds, and the
median over datasets of the per-dataset standard deviation across seeds.
The seed-to-seed spread of a 47-dataset mean is below one AUROC point for
14 of the 18 methods with a reported spread under contamination and for 17
of the 19 on clean data, and below 1.3 for every method but DeepSVDD; the
per-dataset spread does not track dataset size.

\begin{table}[t]
\centering
\caption{Seed variability of every method in Tables~\ref{tab:pdnoisy1}--\ref{tab:pdclean2}: the
47-dataset mean AUROC ($\times100$), the standard deviation of that mean over
the five seeds, and the median over datasets of the per-dataset standard
deviation over seeds; both regimes. Markers as in Table~\ref{tab:main}.
$^{\mathrm{a}}$\,TCCM's contaminated-regime per-seed values were not
retained (Appendix~\ref{app:baselines}); the mean is Table~\ref{tab:main}'s. $^{\dagger}$\,uTabPFN covers 44 of 47 datasets.}
% Source: audit_scripts/seed_std_p140.py over the official per-seed cells (gated to the release).
\label{tab:seedstd}
\medskip
\small
\begin{tabular}{lrrrrrr}
\toprule
 & \multicolumn{3}{c}{Contaminated} & \multicolumn{3}{c}{Clean}\\
\cmidrule(lr){2-4}\cmidrule(lr){5-7}
Method & mean & sd of mean & median sd & mean & sd of mean & median sd\\
\midrule
FOCUS (ours) & 80.0 & 0.96 & 2.9 & 85.9 & 0.51 & 1.2\\
ZEN (ours) & 79.7 & 0.84 & 2.8 & 85.2 & 0.38 & 1.5\\
IForest & 76.4 & 0.44 & 2.3 & 79.6 & 0.46 & 1.8\\
TCCM$^{\mathrm{d}}$ & 72.2$^{\mathrm{a}}$ & -- & -- & 80.8 & 0.69 & 2.0\\
TabPFN-PL$^{\mathrm{t}}$ & 74.6 & 0.45 & 1.8 & 78.8 & 0.17 & 1.3\\
COPOD & 74.4 & 0.35 & 1.7 & 74.5 & 0.47 & 1.0\\
HBOS & 74.4 & 0.23 & 2.2 & 77.4 & 0.57 & 1.3\\
DTE$^{\mathrm{d}}$ & 74.3 & 0.73 & 3.4 & 81.5 & 0.47 & 1.6\\
raw $k$NN & 73.9 & 0.16 & 2.1 & 85.1 & 0.56 & 0.7\\
ECOD & 73.9 & 0.52 & 2.0 & 74.2 & 0.36 & 0.8\\
CBLOF & 73.7 & 0.54 & 2.9 & 81.8 & 0.83 & 1.6\\
PCA & 72.5 & 0.95 & 1.9 & 76.7 & 0.45 & 1.0\\
DeepSVDD$^{\mathrm{d}}$ & 70.9 & 2.31 & 5.5 & 74.7 & 1.12 & 4.1\\
OCSVM & 70.1 & 1.28 & 2.0 & 75.6 & 0.39 & 0.7\\
SOD & 69.1 & 0.80 & 3.4 & 58.7 & 0.96 & 2.0\\
uTabPFN$^{\mathrm{t}}$$^{\dagger}$ & 68.3 & 1.06 & 3.2 & 72.0 & 0.50 & 1.7\\
COF & 66.4 & 0.94 & 3.3 & 55.7 & 0.74 & 2.0\\
LODA & 66.0 & 1.20 & 6.8 & 69.8 & 1.04 & 4.6\\
LOF & 65.4 & 0.65 & 3.2 & 82.9 & 0.58 & 1.1\\
\bottomrule
\end{tabular}
\end{table}

\section{Statistical comparisons}
\label{app:stats}

\subsection{Construction of the main comparison (Figure~\ref{fig:main})}
\label{app:blindcd}
Figure~\ref{fig:main} compares FOCUS with the baselines in the
contaminated regime, following the procedure of
\citet{demsar2006statistical} for comparing one proposed method with
several others. Methods are ranked within each dataset on their
five-seed mean AUROC, rank 1 the best, and the axis of the figure is
the mean rank over the 47 datasets. The ranked pool holds FOCUS and
the 16 baselines that cover all 47 datasets; uTabPFN covers 44.
ZEN, our own frozen detector, is not a baseline; its comparison with
FOCUS is in Sec.~\ref{sec:results}. A Friedman test over the 17
methods rejects the hypothesis that all perform alike ($p<10^{-13}$).
FOCUS is then tested against each of the 16 baselines with a two-sided
Wilcoxon signed-rank test on the 47 paired values
\citep{benavoli2016posthoc}, and the 16 $p$-values are Holm-corrected.
All 16 are significant; the largest Holm-adjusted $p$-value is
$0.034$, against raw-feature $k$NN and TabPFN-PL. The bars among the
baselines are critical-difference cliques: Wilcoxon tests over all 120
of their pairs, Holm-corrected, with a bar joining each maximal run of
adjacent methods among which no pair differs significantly. Against
uTabPFN, on its 44 datasets, FOCUS is ahead by 13.1 AUROC points
($p<0.001$). FOCUS beats Isolation Forest on 28 of the 47 datasets;
the signed-rank test weighs the size of each difference, and FOCUS's
wins are the larger ones.

\subsection{Per-regime diagrams and the pairwise matrix}
\label{app:perregime}
\begin{figure}[t]
\centering
% Source: final/figures/paper_fig_cd_v10.pdf (build_results_v10.py).
\includegraphics[width=0.96\textwidth]{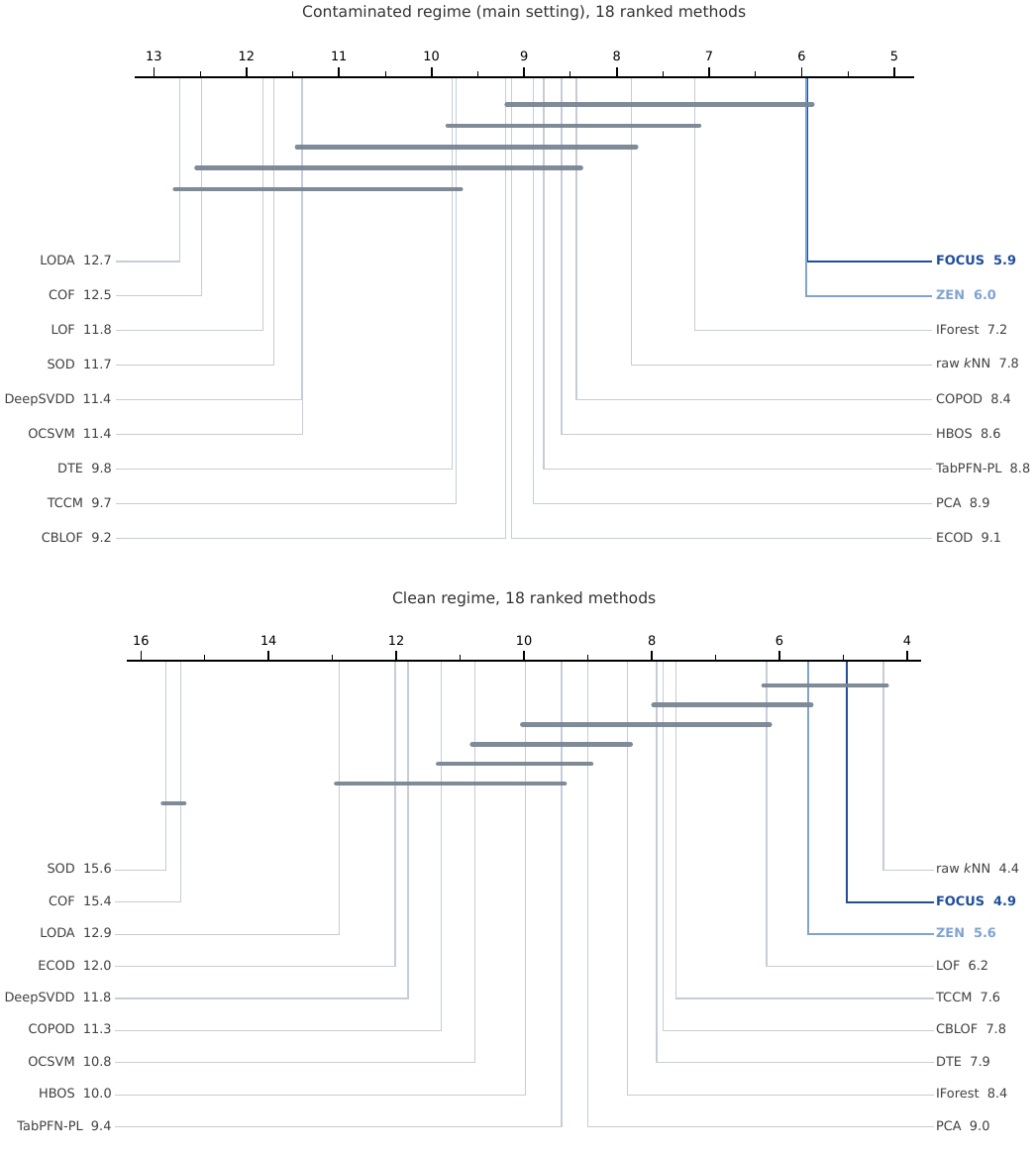}
\caption{Critical-difference diagrams over the 47 datasets, best mean
rank on the right; bars join methods whose pairwise Wilcoxon differences
are not significant after Holm correction over all pairs (153 in
each regime; we use Holm-corrected pairwise tests rather than the
mean-rank post-hoc test, following \citealp{benavoli2016posthoc}).
Top: contaminated regime, 18 ranked methods. Bottom: clean regime, 18
ranked methods. uTabPFN is excluded from the ranks.}
\label{fig:cdapp}
\end{figure}

\begin{figure}[t]
\centering
% Source: final/figures/paper_fig_mcm_v10.pdf (build_results_v10.py).
\includegraphics[width=0.98\textwidth]{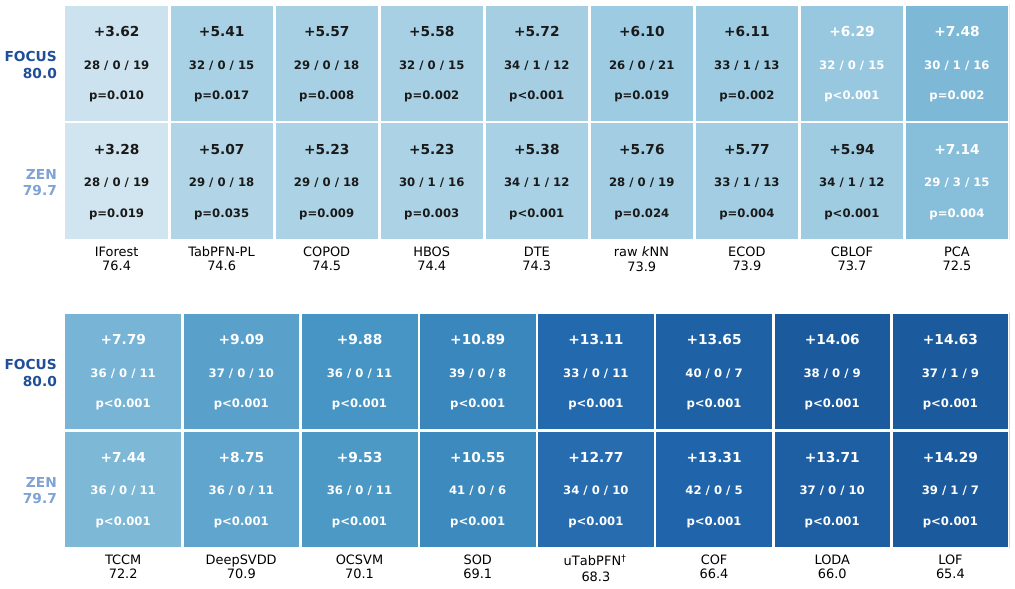}
\caption{Multi-comparison matrix of our methods against the
seventeen baselines, contaminated regime, split into two column
halves, in the format of \citet{ismailfawaz2023mcm} as adopted by recent
large-scale benchmark studies \citep{middlehurst2024bakeoff}. Each cell
gives the mean per-dataset AUROC difference (row minus column), the
win\,/\,tie\,/\,loss count for the row method, and the Wilcoxon
p-value as computed, before any correction, all in bold when $p<0.05$;
cell color encodes the mean
difference (blue: row ahead; orange: column ahead). Rows and columns are
ordered by mean AUROC. $^{\dagger}$uTabPFN covers 44 of 47 datasets.}
\label{fig:mcm}
\end{figure}

Figure~\ref{fig:main} tests FOCUS against the baselines in the
contaminated regime; this subsection ranks every method in each
regime. Because rank-based summaries depend on which
methods are in the ranked pool \citep{benavoli2016posthoc},
Figure~\ref{fig:main} draws its significance from pairwise per-dataset
tests; this subsection reports the all-pairs rank-based comparisons. Figure~\ref{fig:cdapp} gives the Holm-corrected
critical-difference diagram of each regime separately, over every
ranked method including ZEN and FOCUS.
Figure~\ref{fig:mcm} reports the complete pairwise picture for the
contaminated regime: the mean AUROC difference, the win/tie/loss
count, and the Wilcoxon $p$-value of our methods against every
baseline, with each $p$-value reported as computed, before any
multiple-comparison correction.

\subsection{Regret profiles and the remaining gap}
\label{app:ensregret}
\begin{figure}[t]
\centering
% Source: final/figures/paper_fig_regret.pdf (build_results_v10.py).
\includegraphics[width=0.62\textwidth]{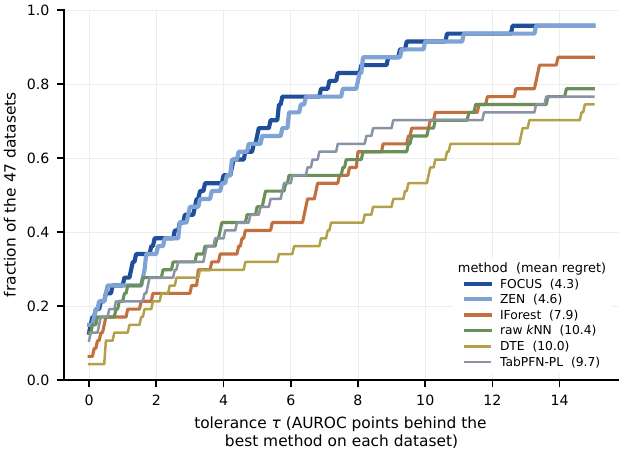}
\caption{Regret profiles, contaminated regime, adapted from the
performance profiles of \citet{dolan2002profiles} with their runtime
ratio replaced by the AUROC gap: the fraction of the 47 datasets on
which a method is within $\tau$ AUROC points of that dataset's best
method. Mean regret in the legend; higher and further left is better.
The per-dataset best is an oracle that changes identity across
datasets; every method trails it somewhere, ours least. Shown: our
methods and the strongest baselines.}
\label{fig:regret}
\end{figure}

Figure~\ref{fig:regret} shows how far each method trails the
per-dataset best across the 47 datasets. On each dataset, a method's
regret is the AUROC of the best compared method on that dataset minus
its own. The profile plots, for every tolerance $\tau$, the fraction of
datasets on which the method's regret is at most $\tau$; a curve that
rises faster belongs to a method that is rarely far from the best.
Compared with an oracle that
picks the best method for each dataset, FOCUS trails by 4.3 AUROC points
on average and ZEN by 4.6, the two smallest gaps in the comparison. The
largest gaps appear in two situations. On shuttle, Cardiotocography,
vertebral and pendigits, four datasets with at most 21 features, Isolation
Forest is well ahead of our detectors. On Wilt, the point above the diagonal in
Figure~\ref{fig:analysis}b, the reference anomalies do not look isolated
to the estimate, so the cleaning misfires and costs 6 AUROC points; over
all 47 datasets the cleaning helps by more than one AUROC point on 27 and
costs more than one point on 10. On Speech no method in the comparison, ours included, reaches 55
AUROC in either regime; vertebral and yeast are the only other such
datasets. Speech's features are nearly
Gaussian, so the challenge there is the representation itself, and
semantic context in the spirit of ReTabAD \citep{yoon2026retabad} is a
promising complement.


\subsection{Ablations: every step of ZEN and FOCUS}
\label{app:ladder}

\begin{figure}[t]
\centering
% Source: final/figures/paper_fig_ftgain_v16.pdf (build_fig4_v16.py, F103), the former Fig. 4(c).
\includegraphics[width=0.42\textwidth]{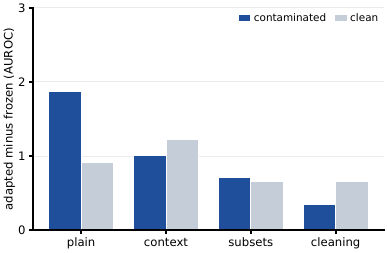}
\caption{The fine-tune's gain at each step of the ladder: the adapted
model minus the frozen model, mean AUROC over the 47 datasets, both
regimes.}
\label{fig:ftgain}
\end{figure}

\begin{figure}[t]
\centering
% Source: final/figures/paper_fig_estimate_v17.pdf (build_fig4_v17.py, F103; v16 numbers, label wording only).
\includegraphics[width=0.42\textwidth]{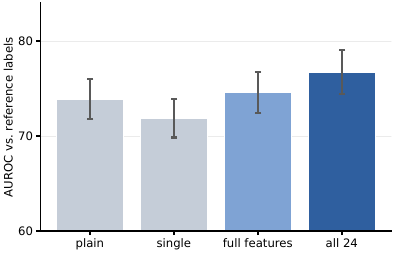}
\caption{The isolation estimate against the reference labels,
contaminated regime: AUROC of the estimate as an anomaly score for the
reference samples, computed from one plain embedding, from one
representation (one feature subset, one block), from the full set of
features pooled over its four blocks, and from all 24 representations
pooled as in ZEN; bars are standard errors.}
\label{fig:estimate}
\end{figure}

Table~\ref{tab:ladder} lists the ladder of Figure~\ref{fig:analysis}a for
both regimes and both models, recomputed on CPU from the official run's
saved embeddings with the official constants; the last step reproduces the
ZEN and FOCUS means of the main runs to within 0.05 AUROC
(half-precision storage). Each step is compared with the previous one by a Wilcoxon
signed-rank test over the 47 dataset means, and Figure~\ref{fig:ftgain}
plots the difference between the two models at each step, the
fine-tune's gain, in both regimes. The extra line of each
contaminated block cleans every one of the 24 representations with its
own weights instead of the pooled ones; Appendix~\ref{app:sens} varies
every constant of the readout on the same runs. The isolation
estimate of Sec.~\ref{sec:zen} ranks the reference samples of each
contaminated run (Figure~\ref{fig:estimate}); against the reference
labels, one plain embedding scores 73.9 AUROC, the average single
representation (one feature subset, one block) 71.9, the full set of
features pooled over its four blocks 74.6, the twenty random-subset
representations pooled 76.2, and all 24 pooled 76.7. The
pooled estimate beats the single representations on 43 of the 47
datasets ($p<10^{-10}$). Among the 5\% of reference samples the pooled
estimate distrusts most, 31\% are anomalies, against 12\% in the
reference set as a whole, and the trust weights built from it average
0.70 on a reference anomaly and 1.16 on a normal sample. The labels enter
this diagnostic only; no method sees them.

\paragraph{What the fine-tune adds.}
The more of ZEN's steps are in place, the less the fine-tune adds.
Figure~\ref{fig:ftgain} reports FOCUS's gain over the frozen model at
each step of the ladder: about 1.9 AUROC points on the plain readout
under contamination, 0.3 after the soft cleaning. One possible reading,
which we have not tested directly, is that the fine-tune and the soft
cleaning repair the same defect, a reference set polluted by anomalies.
On clean data, where there is nothing to clean, the fine-tune keeps a
significant gain of 0.6 to 1.2 AUROC points at every step.

\begin{table}[t]
\centering
\caption{The ablation ladder of Figure~\ref{fig:analysis}a in numbers:
mean AUROC over the 47 datasets at each step, the change from the
previous step with its Wilcoxon $p$-value, and the number of datasets that
improve. The last step of each block is the reported method (Table~\ref{tab:main}'s
clean FOCUS, 85.9, recomputes here as 85.8, within the stated tolerance);
in the clean regime the soft cleaning is switched off ($\lambda=0$), so
its row changes nothing.}
% Body: final/figures/paper_tab_ladder_v10.tex (build_fig4_v10.py, F103).
\label{tab:ladder}
\medskip
\small
\begin{tabular}{lrrrr}
\toprule
Step & AUROC & vs previous & $p$ & wins\\
\midrule
\multicolumn{5}{l}{\textit{Contaminated regime, frozen model (ZEN)}}\\
plain readout & 74.1 & -- & -- & --\\
+ augmented context & 75.0 & +0.9 & 0.130 & 27/47\\
+ feature subsets & 76.9 & +1.9 & $<$0.001 & 38/47\\
+ soft cleaning & 79.7 & +2.8 & $<$0.001 & 36/47\\
\quad soft cleaning per representation, not pooled & 78.6 & -1.1 & $<$0.001 & 10/47\\
\addlinespace[2pt]
\multicolumn{5}{l}{\textit{Contaminated regime, adapted model (FOCUS)}}\\
plain readout & 75.9 & -- & -- & --\\
+ augmented context & 76.0 & +0.1 & 0.216 & 25/47\\
+ feature subsets & 77.6 & +1.6 & $<$0.001 & 34/47\\
+ soft cleaning & 80.0 & +2.5 & $<$0.001 & 35/47\\
\quad soft cleaning per representation, not pooled & 79.2 & -0.9 & $<$0.001 & 12/47\\
\addlinespace[2pt]
\multicolumn{5}{l}{\textit{Clean regime, frozen model (ZEN)}}\\
plain readout & 84.2 & -- & -- & --\\
+ augmented context & 84.3 & +0.1 & 0.698 & 26/47\\
+ feature subsets & 85.2 & +0.9 & 0.003 & 31/47\\
+ soft cleaning (off in this regime) & 85.2 & -- & -- & --\\
\addlinespace[2pt]
\multicolumn{5}{l}{\textit{Clean regime, adapted model (FOCUS)}}\\
plain readout & 85.1 & -- & -- & --\\
+ augmented context & 85.5 & +0.4 & 0.203 & 29/47\\
+ feature subsets & 85.8 & +0.3 & 0.011 & 33/47\\
+ soft cleaning (off in this regime) & 85.8 & -- & -- & --\\
\addlinespace[2pt]
\bottomrule
\end{tabular}
\end{table}

\subsection{Ablation: the readout window}
\label{app:window}
\begin{figure}[t]
\centering
% Source: final/figures/paper_fig_window_p148.pdf (final/scripts/f109_f115/build_ablations_p148.py, F111).
\includegraphics[width=\textwidth]{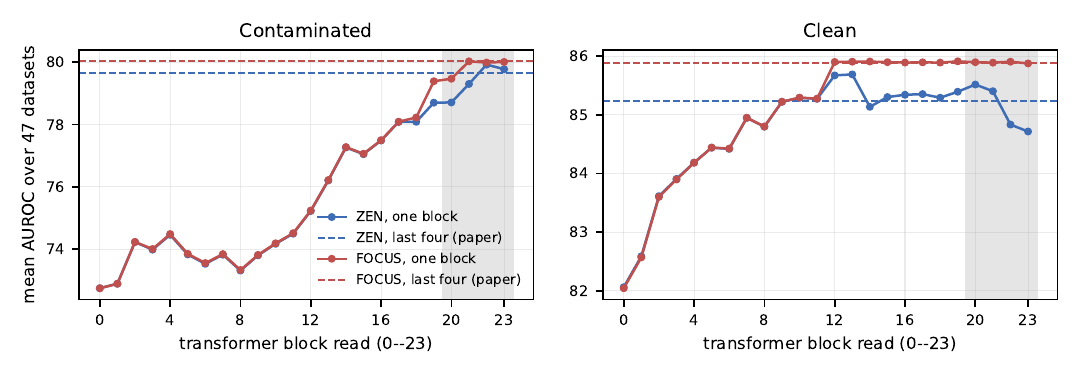}
\caption{Mean AUROC over the 47 datasets, five seeds, when the readout
uses one transformer block at a time (dots), for the frozen model (ZEN)
and the adapted one (FOCUS); dashed lines: the paper's four-block
window, shaded.}
\label{fig:window}
\end{figure}

The readout takes the embeddings of the last four of TabPFN-3's 24
transformer blocks (Sec.~\ref{sec:zen}). Table~\ref{tab:window} and
Figure~\ref{fig:window} vary this choice on the same datasets, seeds and
splits as the main results, in both regimes, for the frozen model (ZEN) and the adapted one
(FOCUS): the last $n$ blocks for $n$ from 1 to 24, three four-block
windows deeper in the network, and, in the figure, every single block.
Every other constant is the paper's, and the adapted model is the
saved fine-tuned model of each run of the main evaluation, so no
fine-tune is rerun.
The four-block window reproduces the official means to within 0.05
AUROC in every case. Under contamination the last one to four blocks are
equivalent, within 0.3 AUROC and never significantly different, and
every wider or deeper window is worse; reading all 24 blocks costs 2.7
points for ZEN and 2.9 for FOCUS, both significant, and single blocks
improve steadily with depth from block 8 on. On clean data the picture
is flatter: the frozen readout gains 0.1 to 0.2 points from wider
windows (last 6 to 12 blocks, $p$ between 0.02 and 0.04) and loses 0.2
to 0.5 from narrower ones, while the adapted model is within 0.05 AUROC
of its reported value for every window from one to twelve blocks. The
last-four window is therefore not a tuned optimum: under contamination
it ties the best windows, and on clean data the best window is 0.3
points away.

\begin{table}[t]
\centering
\caption{The readout window: mean AUROC over the 47 datasets, five
seeds, for the frozen model (ZEN) and the adapted one (FOCUS) in both
regimes, with the difference from the paper's window and its Wilcoxon
$p$-value. Blocks are numbered 0--23; the paper reads blocks 20--23.}
\label{tab:window}
\medskip
\scriptsize
\setlength{\tabcolsep}{3pt}
% Body: final/figures/paper_tab_window_p148.tex (build_ablations_p148.py, F111).
\begin{tabular}{lllll}
\toprule
Readout window & \multicolumn{2}{c}{Contaminated} & \multicolumn{2}{c}{Clean}\\
\cmidrule(lr){2-3}\cmidrule(lr){4-5}
 & ZEN & FOCUS & ZEN & FOCUS\\
\midrule
last block & 79.8 (+0.1, 0.65) & 80.0 ($\pm$0.0, 0.60) & 84.7 (-0.5, $<$0.001) & 85.9 ($\pm$0.0, 0.06)\\
last 2 blocks & 79.9 (+0.3, 0.91) & 80.0 ($\pm$0.0, 0.50) & 84.8 (-0.4, $<$0.001) & 85.9 ($\pm$0.0, 0.06)\\
last 3 blocks & 79.8 (+0.2, 0.54) & 80.0 ($\pm$0.0, 0.49) & 85.0 (-0.2, $<$0.001) & 85.9 ($\pm$0.0, 0.13)\\
last 4 blocks (paper) & 79.7 & 80.0 & 85.2 & 85.9\\
last 6 blocks & 79.3 (-0.4, 0.22) & 79.7 (-0.3, 0.37) & 85.3 (+0.1, 0.04) & 85.9 ($\pm$0.0, 0.48)\\
last 8 blocks & 79.0 (-0.6, 0.19) & 79.4 (-0.7, 0.36) & 85.3 (+0.1, 0.03) & 85.9 ($\pm$0.0, 0.03)\\
last 12 blocks & 78.5 (-1.2, 0.18) & 78.7 (-1.4, 0.33) & 85.5 (+0.2, 0.02) & 85.9 ($\pm$0.0, 0.54)\\
all 24 blocks & 77.0 (-2.7, 0.04) & 77.1 (-2.9, 0.02) & 85.1 (-0.1, 0.46) & 85.5 (-0.4, 0.09)\\
\addlinespace[2pt]
blocks 16--19 & 78.2 (-1.5, 0.07) & 78.4 (-1.6, 0.10) & 85.4 (+0.1, 0.07) & 85.9 ($\pm$0.0, 0.21)\\
blocks 12--15 & 76.6 (-3.1, 0.02) & 76.6 (-3.4, 0.02) & 85.5 (+0.3, 0.03) & 85.9 ($\pm$0.0, 0.45)\\
blocks 8--11 & 74.0 (-5.7, 0.01) & 74.0 (-6.1, 0.004) & 85.2 ($\pm$0.0, 0.54) & 85.2 (-0.7, 0.06)\\
\bottomrule
\end{tabular}
\end{table}

\subsection{Sensitivity of the readout}
\label{app:sens}
Table~\ref{tab:sens} varies each constant of the readout while every
other constant stays at the paper's value, on the saved embeddings of
the main runs in both regimes, five seeds, for the frozen and the
adapted model: the neighbor count $k$, the soft-cleaning strength
$\lambda$, the mixing weight $\alpha$, where the score is $\alpha$ times
the random-subset average plus $1-\alpha$ times the full-feature score,
so that $\alpha=0.5$ is Eq.~\ref{eq:zen} up to a constant factor, and
the number $B$ of random feature subsets. The paper's setting reproduces
the official means to within 0.05 AUROC. Under contamination $k=50$
sits on a plateau: $k=100$ and $200$ are within 0.2 points (not
significant) and smaller $k$ loses up to 2.8. On clean data every $k$
from 5 to 200 stays within 0.8 points of the paper's value, the smaller
values marginally above it; one $k$ serves both regimes. Every $\lambda$ from
0.25 to 2 improves on no cleaning under contamination and stays above
Isolation Forest, by 2.9 to 3.3 points for ZEN, and 0.5 and 1 give the
same mean, so the paper's value sits on a plateau rather than a peak; on
clean data every $\lambda>0$ costs 0.4 to 2.5 points, which is why the
cleaning is switched off there. The equal mixing weight is best in both
regimes, but $\alpha$ between 0.25 and 0.75 stays within 0.3 points,
while the full set of features alone or the random subsets alone cost
0.3 to 1.4. Under contamination each added random subset helps,
significantly up to the fourth; the fifth adds 0.2 points (not
significant), and three random subsets already place ZEN above every
baseline in Table~\ref{tab:main}. On clean data the first random subset
alone is no better than the full set of features alone, and from the
second subset on the average helps.

\begin{table}[t]
\centering
\caption{Sensitivity of the readout: mean AUROC over the 47 datasets,
five seeds, for the frozen model (ZEN) and the adapted one (FOCUS) in
both regimes, one constant varied at a time with the difference from the
paper's value and its Wilcoxon $p$-value. With $B$ random subsets the
trust weight pools the $(B{+}1)\times4$ representations, as ZEN does
with six.}
\label{tab:sens}
\medskip
\scriptsize
\setlength{\tabcolsep}{3pt}
% Body: final/figures/paper_tab_readout_p148.tex (build_ablations_p148.py, F112).
\begin{tabular}{llll l}
\toprule
 & \multicolumn{2}{c}{Contaminated} & \multicolumn{2}{c}{Clean}\\
\cmidrule(lr){2-3}\cmidrule(lr){4-5}
 & ZEN & FOCUS & ZEN & FOCUS\\
\midrule
\multicolumn{5}{l}{\textit{Neighbor count $k$}}\\
5 & 76.9 (-2.8, $<$0.001) & 77.9 (-2.2, 0.002) & 86.0 (+0.8, 0.03) & 86.2 (+0.3, 0.35)\\
10 & 78.0 (-1.7, 0.001) & 78.8 (-1.2, 0.02) & 85.8 (+0.6, 0.01) & 86.2 (+0.3, 0.10)\\
20 & 78.8 (-0.9, 0.004) & 79.5 (-0.5, 0.06) & 85.5 (+0.3, 0.007) & 86.1 (+0.2, 0.05)\\
50 (paper) & 79.7 & 80.0 & 85.2 & 85.8\\
100 & 79.9 (+0.2, 0.45) & 80.1 (+0.1, 0.83) & 85.0 (-0.2, 0.004) & 85.8 (-0.1, 0.11)\\
200 & 79.7 (+0.1, 0.96) & 79.6 (-0.4, 0.15) & 84.8 (-0.4, 0.009) & 85.8 (-0.1, 0.12)\\
\midrule
\multicolumn{5}{l}{\textit{Soft-cleaning strength $\lambda$}}\\
0 (paper, clean) & 76.9 (-2.8, $<$0.001) & 77.6 (-2.5, $<$0.001) & 85.2 & 85.8\\
0.25 & 79.3 (-0.4, 0.001) & 79.7 (-0.3, 0.008) & 84.8 (-0.4, 0.10) & 84.6 (-1.2, 0.02)\\
0.5 (paper, contam.) & 79.7 & 80.0 & 84.5 (-0.7, 0.04) & 84.1 (-1.7, 0.005)\\
1 & 79.7 ($\pm$0.0, 0.76) & 80.1 ($\pm$0.0, 0.81) & 84.0 (-1.1, 0.01) & 83.7 (-2.1, 0.001)\\
2 & 79.3 (-0.4, 0.42) & 79.4 (-0.6, 0.58) & 83.4 (-1.8, $<$0.001) & 83.4 (-2.5, $<$0.001)\\
\midrule
\multicolumn{5}{l}{\textit{Weight of the random subsets $\alpha$}}\\
0 (full set only) & 79.1 (-0.6, 0.002) & 79.5 (-0.6, 0.009) & 84.3 (-0.9, 0.003) & 85.5 (-0.3, 0.01)\\
0.25 & 79.5 (-0.2, 0.02) & 79.9 (-0.2, 0.10) & 84.8 (-0.3, 0.01) & 85.8 ($\pm$0.0, 0.19)\\
0.5 (paper) & 79.7 & 80.0 & 85.2 & 85.8\\
0.75 & 79.6 (-0.1, 0.81) & 80.0 (-0.1, 0.97) & 85.0 (-0.2, 0.42) & 85.5 (-0.3, 0.04)\\
1 (random subsets only) & 79.2 (-0.5, 0.05) & 79.5 (-0.6, 0.02) & 84.0 (-1.2, 0.005) & 84.5 (-1.4, $<$0.001)\\
\midrule
\multicolumn{5}{l}{\textit{Number of random feature subsets $B$}}\\
0 & 77.5 (-2.2, $<$0.001) & 77.9 (-2.1, $<$0.001) & 84.3 (-0.9, 0.003) & 85.5 (-0.3, 0.01)\\
1 & 77.9 (-1.8, $<$0.001) & 78.4 (-1.6, $<$0.001) & 84.1 (-1.1, $<$0.001) & 85.1 (-0.7, $<$0.001)\\
2 & 78.6 (-1.1, $<$0.001) & 79.0 (-1.0, $<$0.001) & 84.8 (-0.4, 0.005) & 85.6 (-0.3, 0.004)\\
3 & 79.2 (-0.4, 0.04) & 79.6 (-0.4, 0.04) & 85.0 (-0.2, 0.02) & 85.7 (-0.1, 0.02)\\
4 & 79.5 (-0.2, 0.59) & 79.8 (-0.2, 0.34) & 85.1 (-0.1, 0.14) & 85.8 (-0.1, 0.11)\\
5 (paper) & 79.7 & 80.0 & 85.2 & 85.8\\
\bottomrule
\end{tabular}
\end{table}

\subsection{Ablations of the fine-tune}
\label{app:ftabl}
Table~\ref{tab:ftabl} varies the constants of the fine-tune: the number
$K$ of trainable blocks, the trust weights of Eq.~\ref{eq:focus}, the
number of epochs, the checkpoint rule and the learning rate. These arms
rerun the fine-tune, so they are run at one seed (seed 0) on every
dataset, on the mixed GPU types of our cluster's spare nodes. The frozen
readout itself differs between GPU types by up to a few AUROC points on
the smallest datasets, so the paper's own setting is rerun under the
same conditions and every arm is compared with that rerun, paired over
the 47 datasets; the rerun reproduces the official seed-0 means to
within 0.05 AUROC in both regimes. Under contamination $K=3$ and the
paper's $K=6$ are equivalent ($-0.5$, $p=0.91$) and deeper fine-tunes
are worse, by 2.6 points at $K=12$ and 3.2 at $K=24$, both significant.
On clean data the paper's $K=12$ is best, $K=3$ is within 0.5 points
(not significant), $K=6$ costs 1.0 ($p=0.04$) and $K=24$ costs 3.0. A
single $K=3$ would therefore serve both regimes within 0.5 AUROC of the
reported values. Switching the trust weights off costs 0.2 AUROC in
both regimes ($p=0.06$ and $0.02$). Under contamination one or three
epochs, or keeping the last epoch's weights instead of the average,
change the mean by at most 0.1 points (not significant); a learning rate
ten times smaller costs 0.4 (not significant) and ten times larger
costs 1.8 ($p=0.05$). On clean data one or three epochs, or keeping the last epoch's weights, change the mean by at most 0.2 points (ep3: $p=0.03$); a learning rate ten times smaller changes it by +0.1 ($p=0.05$) while a rate ten times larger destabilizes the fine-tune ($-19.9$ points, $p<0.001$): lower rates than the paper's are safe, higher ones are not.

\begin{table}[t]
\centering
\caption{Ablations of the fine-tune, seed 0, 47 datasets, mixed GPU
types: mean AUROC of FOCUS with the difference from the paper's
setting rerun under the same conditions and its Wilcoxon $p$-value.
Every arm keeps the paper's other constants; the last row is the frozen
model on the same runs.}
\label{tab:ftabl}
\medskip
\small
% Body: final/figures/paper_tab_ftabl_p148.tex (build_ablations_p148.py, F109 + F113).
\begin{tabular}{lll}
\toprule
Fine-tuning variant & Contaminated & Clean\\
\midrule
\multicolumn{3}{l}{\textit{Trainable blocks $K$ (paper: 6 contaminated, 12 clean)}}\\
$K=3$ & 80.5 (-0.5, 0.91) & 85.5 (-0.5, 0.14)\\
$K=6$ & 81.0 (paper) & 85.0 (-1.0, 0.04)\\
$K=12$ & 78.3 (-2.6, 0.004) & 86.0 (paper)\\
$K=24$ & 77.8 (-3.2, 0.002) & 83.1 (-3.0, $<$0.001)\\
\addlinespace[2pt]
\multicolumn{3}{l}{\textit{Other constants, at the paper's $K$}}\\
trust weights off (plain compactness loss) & 80.8 (-0.2, 0.06) & 85.8 (-0.2, 0.02)\\
1 epoch & 80.9 ($\pm$0.0, 0.60) & 85.9 (-0.2, 0.37)\\
3 epochs (three checkpoints averaged) & 81.1 (+0.1, 0.66) & 85.8 (-0.2, 0.03)\\
last epoch kept instead of the average & 80.9 ($\pm$0.0, 0.95) & 86.0 (-0.1, 0.68)\\
learning rate $\times 0.1$ & 80.6 (-0.4, 0.45) & 86.2 (+0.1, 0.05)\\
learning rate $\times 10$ & 79.2 (-1.8, 0.05) & 66.1 (-19.9, $<$0.001)\\
\addlinespace[2pt]
ZEN on the same runs & 80.0 & 85.1\\
\bottomrule
\end{tabular}
\end{table}

\subsection{Ablation: the context of each run}
\label{app:context}
Table~\ref{tab:context} varies the number $m$ of synthetic context
samples and the number of leave-one-fold-out folds of
Sec.~\ref{sec:zen} for the frozen model under contamination, at seed 0
on every dataset (the arms re-embed the reference set, on mixed GPU
types); the paper's setting recomputed on the same runs is the
reference, and it reproduces the official seed-0 mean to within 0.2
AUROC. Without synthetic samples ZEN loses 0.4 points and with 100 of
them 0.7 ($p=0.18$ and $0.06$); with up to 2,000 it gains 0.8 (not
significant; on datasets with at most 500 reference samples this arm
equals the paper's). The augmented context is thus a small, consistent
component. Five or two folds instead of
ten change the mean by $+0.6$ and $+0.5$ points, neither significant:
the reference set can be embedded in two passes instead of ten at no
measurable cost, the remedy Sec.~\ref{sec:limitations} names.

\begin{table}[t]
\centering
\caption{The context of each run, contaminated regime, seed 0, 47
datasets: mean AUROC of ZEN with the difference from the paper's
setting and its Wilcoxon $p$-value.}
\label{tab:context}
\medskip
\small
% Body: final/figures/paper_tab_context_p148.tex (build_ablations_p148.py, F114).
\begin{tabular}{ll}
\toprule
Context construction & ZEN, contaminated\\
\midrule
no synthetic samples ($m=0$) & 79.4 (-0.4, 0.18)\\
$m=100$ & 79.2 (-0.7, 0.06)\\
$m=\min(n,500)$, 10 folds (paper) & 79.9\\
$m=\min(n,2000)$ & 80.6 (+0.8, 0.56)\\
5 folds & 80.5 (+0.6, 0.18)\\
2 folds & 80.4 (+0.5, 0.92)\\
\bottomrule
\end{tabular}
\end{table}

\subsection{Injected contamination}
\label{app:inject}
\begin{figure}[t]
\centering
% Source: final/figures/paper_fig_inject_p148.pdf (build_ablations_p148.py, F115).
\includegraphics[width=\textwidth]{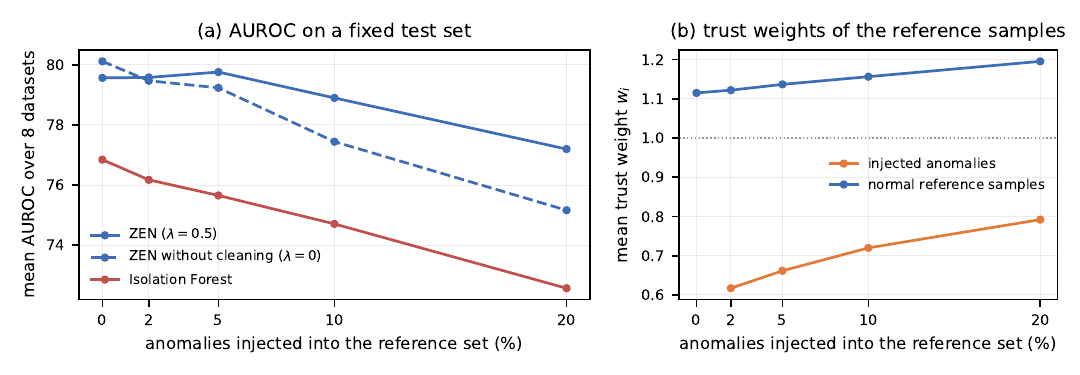}
\caption{Injected contamination on eight datasets, five seeds. (a) Mean
AUROC on a test set that is identical at every rate, as anomalies are
injected into the clean reference set; ZEN with and without the soft
cleaning, and Isolation Forest. (b) Mean trust weight of the
injected anomalies and of the normal reference samples.}
\label{fig:inject}
\end{figure}

The contaminated regime of the main experiments carries each dataset's
natural anomaly rate. Figure~\ref{fig:inject} controls the rate
instead, on the eight datasets whose anomaly rate is at least 30\%
(breastw, fault, Ionosphere, magic.gamma, Pima, satellite, SpamBase and
yeast), the ones on which a 20\% reference contamination is feasible
while half of the anomalies stay in the test set. The reference set is
the clean regime's, normal samples only; half of the test anomalies,
drawn once per seed, form an injection pool, and the first
$\mathrm{round}\bigl(\tfrac{r}{1-r}\,n\bigr)$ of them are appended to
the $n$ reference normals for the rate $r$ in $\{0, 2, 5, 10, 20\%\}$,
so that the injected sets are nested and the test set, the clean test
normals plus the other half of the anomalies, is identical at every
rate. The feature scaler is fitted once on the clean reference set.
We run ZEN with the contaminated readout ($\lambda=0.5$), the same
embeddings read without cleaning ($\lambda=0$), and the strongest
contaminated-regime baseline of Table~\ref{tab:main}, Isolation Forest,
on five seeds; FOCUS is not rerun per rate. ZEN's gain over its
own uncleaned readout grows with the rate, from 0.1 points at 2\% to
0.5, 1.5 and 2.0 at 5, 10 and 20\% ($p=0.08$ at 20\% on eight
datasets), and its margin over Isolation Forest grows from 2.7 to 4.6
points, significant from 5\% on. On the
clean reference set the cleaning costs 0.5 points (not significant),
consistent with $\lambda=0$ in the clean regime. Panel (b) shows the
mechanism: the trust weight of the injected anomalies averages 0.62 to
0.79 against 1.12 to 1.20 for the normal reference samples, and the
isolation estimate ranks the injected samples with 75 to 79 AUROC
against the injection labels. SpamBase's clean
test set keeps four normal samples (Appendix~\ref{app:splits}), so its
curve is the noisiest of the eight.

\subsection{ZEN on other backbones}
\label{app:backbones}
\begin{figure}[t]
\centering
% Source: final/figures/paper_fig_backbones_v2.pdf (build_backbones_v2.py).
\includegraphics[width=0.8\textwidth]{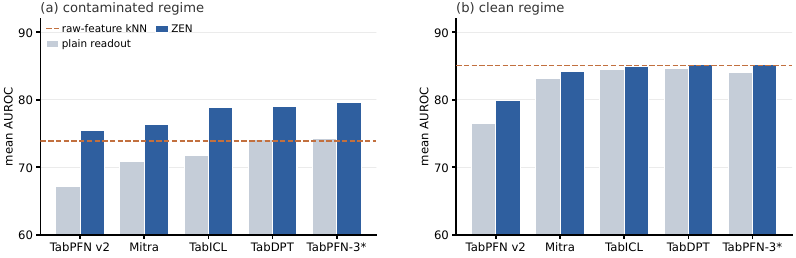}
\caption{Plain readout and ZEN on five in-context backbones, both
regimes; dashed line: raw-feature $k$NN; the asterisk marks the paper's
backbone, TabPFN-3. Figure~\ref{fig:analysis}c is panel (a) without the
raw-feature $k$NN line.}
\label{fig:backbones}
\end{figure}

Table~\ref{tab:backbones} and Figure~\ref{fig:backbones} report the plain
readout and ZEN on five backbones in both regimes: TabPFN-3, the paper's
backbone; TabPFN v2 \citep{hollmann2025tabpfnv2}; TabICL
\citep{qu2025tabicl}, in its TabICLv2 checkpoint \citep{qu2026tabiclv2};
TabDPT \citep{ma2025tabdpt}; and Mitra \citep{zhang2025mitra}. Every
constant of Sec.~\ref{sec:zen} and Appendix~\ref{app:config} is
unchanged, and every model runs frozen with the full reference set as its
context, a single estimator and its test-time augmentations switched off.
The one rule applied per backbone is where the embeddings are taken:
the last four blocks of the model's in-context transformer, at the point
where the model itself uses its last block. These are blocks 20--23 of 24 for TabPFN-3, 8--11 of 12
for TabPFN v2, TabICL and Mitra, and 28--31 of 32 for TabDPT. Each
block's output passes through the model's own final normalization where
the model has one; TabDPT's running representation is taken directly,
and Mitra's at the label slot of each query sample. We check on every
run that the captured last block reproduces the model's own output, to
within 0.02 in predicted probability.
TabDPT accepts at most 128 features, so on the five datasets with more
features each run uses a fixed random subset of 128 of them, the same
device as the paper's 200-feature cap. Every run used the same GPU
model and the paper's five seeds. The TabPFN-3 arm reproduces the
paper's ZEN on every dataset to within 0.5 AUROC, and raw-feature $k$NN,
which needs no forward pass, reproduces the paper's values exactly.

Under contamination ZEN's steps help on every backbone, with the largest
gain on the weakest one. In the clean regime the soft cleaning is
switched off and only the feature subsets and the augmented context act; the
gain is then smaller, significant for TabPFN-3, TabPFN v2, TabICL and
Mitra, and not for TabDPT (0.6 AUROC points, $p=0.36$). Between
backbones, TabPFN-3 leads TabPFN v2 significantly in both regimes (by 4.2
and 5.3 AUROC points); its leads over TabICL (0.8, $p=0.13$; 0.3,
$p=0.88$) and over TabDPT (0.7, $p=0.26$; 0.0, $p=0.78$) are not
significant. Its lead over Mitra is significant under contamination
(3.4, $p=0.006$) and not on clean data (1.0, $p=0.24$); Mitra in turn
leads TabPFN v2 significantly on clean data (4.3, $p<0.001$) but not
under contamination (0.8, $p=0.28$). On clean data TabPFN-3 (85.2) and
TabDPT (85.2) reach raw-feature $k$NN (85.1), TabICL (84.9) and Mitra
(84.2) stay just below it, and TabPFN v2 (79.9) well below.

\paragraph{FOCUS on TabPFN v2.}
The fine-tune was also run on TabPFN v2 under contamination, with the
contaminated regime's constants unchanged ($K=6$ of its 12 blocks) and
the paper's five seeds. Three datasets (backdoor, census and speech)
exceed the memory of our largest GPU on this arm and are left out, so
the arm covers 44 datasets. There the fine-tune lifts the plain readout
from 67.6 to 70.7 AUROC ($+3.0$, $p<0.001$, 31 of 44 datasets), against
$+2.0$ ($p=0.002$) on TabPFN-3 in the same run, and FOCUS reaches 76.1 AUROC, above
ZEN's 75.9 on the same datasets: as on TabPFN-3, the fine-tune's gain
is large on the plain readout and small once ZEN's steps are in place
(Sec.~\ref{sec:analysis}). This arm ran on other GPU types than the
runs of Table~\ref{tab:backbones}; the TabPFN-3 arm of the same run, on
the paper's GPU type, reproduces the paper's FOCUS to within 0.05
AUROC.

\begin{table}[t]
\centering
\caption{ZEN on five backbones: mean AUROC of the plain readout and of
ZEN over the 47 datasets, ZEN's gain with its Wilcoxon $p$-value and the
number of datasets that improve, and the number of datasets on which ZEN
beats raw-feature $k$NN. Every constant is the paper's.}
% Body: final/figures/paper_tab_backbones_v2.tex (build_backbones_v2.py, F104/F106).
\label{tab:backbones}
\medskip
\small
\begin{tabular}{lrrrlr}
\toprule
Backbone & blocks & plain & ZEN & gain ($p$, wins) & ZEN $>$ raw $k$NN\\
\midrule
\multicolumn{6}{l}{\textit{Contaminated regime}}\\
TabPFN v2 & 12 & 67.2 & 75.4 & +8.2 ($<$0.001, 43/47) & 22/47\\
Mitra & 12 & 70.9 & 76.3 & +5.4 ($<$0.001, 37/47) & 22/47\\
TabICL & 12 & 71.7 & 78.9 & +7.2 ($<$0.001, 39/47) & 26/47\\
TabDPT & 32 & 74.2 & 79.0 & +4.9 ($<$0.001, 36/47) & 29/47\\
TabPFN-3 & 24 & 74.2 & 79.7 & +5.5 ($<$0.001, 36/47) & 28/47\\
raw-feature $k$NN (no backbone) & -- & \multicolumn{2}{c}{73.9} & -- & --\\
\addlinespace[2pt]
\multicolumn{6}{l}{\textit{Clean regime}}\\
TabPFN v2 & 12 & 76.4 & 79.9 & +3.4 ($<$0.001, 38/47) & 11/47\\
Mitra & 12 & 83.1 & 84.2 & +1.1 (0.003, 33/47) & 20/47\\
TabICL & 12 & 84.5 & 84.9 & +0.4 (0.045, 31/47) & 24/47\\
TabDPT & 32 & 84.6 & 85.2 & +0.6 (0.356, 24/47) & 22/47\\
TabPFN-3 & 24 & 84.1 & 85.2 & +1.1 (0.010, 28/47) & 21/47\\
raw-feature $k$NN (no backbone) & -- & \multicolumn{2}{c}{85.1} & -- & --\\
\addlinespace[2pt]
\bottomrule
\end{tabular}
\end{table}

\section{Implementation details}
\label{app:impl}
\subsection{Method configuration}
\label{app:config}
Table~\ref{tab:config} lists every constant of ZEN and FOCUS.
Everything not marked per-regime is shared. Nothing in the table is
ever chosen per dataset.

\paragraph{The PFN pretraining objective.}
With $D_{\mathrm{train}}$, $x_{\mathrm{test}}$ and $q_{\theta}$ as in
Sec.~\ref{sec:pfn}, the parameters of a PFN minimize
\begin{equation}
\label{eq:pfnloss}
\mathcal{L}_{\mathrm{PFN}}(\theta)\;=\;
\mathbb{E}_{D\sim p(D)}\bigl[-\log
q_{\theta}(y_{\mathrm{test}}\mid x_{\mathrm{test}},D_{\mathrm{train}})\bigr],
\end{equation}
where each synthetic dataset $D$ drawn from the prior is split into a
context $D_{\mathrm{train}}$ and held-out test pairs
$(x_{\mathrm{test}},y_{\mathrm{test}})$ \citep{hollmann2023tabpfn}.

\begin{table}[t]
\centering
\caption{Complete configuration of our methods. The upper block is
shared by both regimes; the lower block lists the per-regime
constants.}
% Values: anon-release configs focus_{clean,contaminated}.json (F102).
\label{tab:config}
\medskip
\small
\begin{tabular}{ll}
\toprule
Backbone & TabPFN-3 \citep{grinsztajn2026tabpfn3}, 24 transformer blocks\\
 & (\texttt{tabpfn} package 8.1.0, default checkpoint)\\
Readout blocks & the last four of the 24\\
Feature subsets & five random subsets of size $\max(2,\mathrm{round}(d/2))$\\
 & and the full set of features, drawn once per dataset and seed\\
Combination & mean of the random subsets $+$ full set, equal weight (Eq.~\ref{eq:zen})\\
Synthetic context samples & $m=\min(n,500)$ per feature subset, uniform on $[-1,1]$,\\
 & arbitrary labels, never scored\\
Neighbor count $k$ & 50\\
Soft-cleaning weight & $w_i=\exp(-\lambda\hat a_i)$, $\hat a_i$ pooled over the 24 (subset, block)\\
 & representations (Sec.~\ref{sec:zen}), clipped to $[10^{-3},10^{3}]$\\
Score standardization & per block; $\mu,\sigma$ from the reference set's weighted\\
 & leave-one-out $k$NN distances\\
Embedding extraction & leave-one-fold-out, 10 folds, 1 estimator\\
Context labels & fixed arbitrary (random 0/1, drawn per seed)\\
Trust weights (FOCUS) & same rule on the frozen final-block embedding of the\\
 & reference set alone (all features, no synthetic samples),\\
 & $k{=}20$, $\lambda{=}0.5$ in both regimes\\
Center $c$ & trust-weighted mean of the frozen final-block embeddings\\
Loss (FOCUS) & trust-weighted squared distance to $c$ (Eq.~\ref{eq:focus})\\
Epochs & 2, no validation split, no early stopping\\
Kept weights & mean of the epoch-1 and epoch-2 parameters\\
Optimizer & AdamW, constant learning rate\\
Wide inputs & 200-feature cap on three datasets (App.~\ref{app:scope})\\
\bottomrule
\end{tabular}

\medskip

\begin{tabular}{lcc}
\toprule
 & clean & contaminated\\
\midrule
Soft-cleaning strength $\lambda$ & 0 & 0.5\\
Learning rate & $10^{-3}$ & $10^{-4}$\\
Batch size & 32 & 16\\
Trainable blocks $K$ & 12 & 6\\
Gradient clipping & none & 0.01\\
\bottomrule
\end{tabular}
\end{table}

\paragraph{Embedding extraction.}
TabPFN expects labeled context samples and anomaly detection provides
none, so each run passes fixed arbitrary labels: a random 0/1 vector
drawn once from the run's seed. The labels carry no information about
which samples are anomalous, and the readout never uses the classifier
head. Reference-set embeddings are extracted leave-one-fold-out with
ten folds, following \citet{ye2025closerlook}: each fold is embedded
in the query role against the remaining folds, so training and test
embeddings come from the same code path in the same role. When the
reference set holds fewer than ten samples, the fold count drops to
that number, never below two. Each extraction uses a single estimator
for determinism.

\paragraph{Feature subsets and synthetic context samples.}
The $B$ random feature subsets hold $\max(2,\mathrm{round}(d/2))$
features each, drawn without replacement once per dataset and seed.
Random feature subsets are the device behind random forests
\citep{breiman2001random} and feature-bagging outlier ensembles
\citep{lazarevic2005feature}, and the motivation is the same here: runs
that make different mistakes average out one another's errors
(Sec.~\ref{sec:related}). A run's context is its feature subset of
the reference set plus $m=\min(n,500)$ synthetic samples drawn
uniformly on $[-1,1]$ over its features, with the same kind of arbitrary
labels as the reference samples. The synthetic samples are never scored
and never serve as neighbors; they only change what the model
conditions on when it embeds the real samples, and they stay in every
fold's context. Figure~\ref{fig:analysis}a, Appendix~\ref{app:ladder}
and Appendix~\ref{app:context} show their effect.

\paragraph{The FOCUS objective.}
\label{app:focusdetails}
Each reference sample $x_i$ carries a trust weight $w_i$ of the form of
Sec.~\ref{sec:zen}, computed on the frozen final-block embedding of the
reference set alone, all features and no synthetic samples, with its own
constants, $k{=}20$ and $\lambda{=}0.5$, the same in both regimes (the
readout's $\lambda$ is 0 on clean data, the fine-tune's is not). The
same weights define the center as the weighted mean of the frozen
embeddings and enter the loss that FOCUS minimizes over batches
$\mathcal{B}$ of reference samples:
\begin{equation}
\label{eq:focus}
\mathcal{L}(\theta)\;=\;\frac{\sum_{x_i\in\mathcal{B}} w_i\,
\bigl\lVert\psi_{\theta}(x_i\mid D_{\mathrm{train}})-c\bigr\rVert_2^{2}}
{\sum_{x_i\in\mathcal{B}} w_i},
\qquad
c\;=\;\frac{\sum_i w_i\,\psi_{0}(x_i\mid D_{\mathrm{train}})}{\sum_i w_i}.
\end{equation}
Here $\psi_{\theta}$ is the extractor once its weights update and,
without a block index, denotes the final-block embedding; the sums
defining $c$ run over the whole reference set. Weights and center are
computed once from the frozen model and never updated: recomputing the
weights as the model adapts would let it down-weight whatever it fails
to contract, and a fixed center avoids the collapse failure mode of
learned centers \citep{ruff2018deepsvdd}. Normalizing by the weight sum
keeps the loss scale, and with it the effective step size, independent
of the trust strength. With all weights equal, Eq.~\ref{eq:focus} is the
plain compactness loss; with the trust weights it is what we call
\emph{trust-weighted compactness fine-tuning}. FOCUS updates only the
last $K$ blocks of TabPFN and trains on the reference set alone,
embedded leave-one-fold-out with the folds above, so it compacts a set
that is also the model's context; in the contaminated regime that set
contains anomalies, and FOCUS is never told which samples they are.

\paragraph{Fine-tuning.}
FOCUS updates the last $K$ of TabPFN's 24 transformer blocks, $K{=}12$
in the clean regime and $K{=}6$ in the contaminated one, with the
learning rate, batch size and gradient clipping of
Table~\ref{tab:config}. Training runs for two epochs, and the kept
model is the mean of the epoch-1 and epoch-2 parameters.
Appendix~\ref{app:ftabl} ablates $K$, the trust weights, the epochs, the
checkpoint rule and the learning rate.

\paragraph{How the constants were chosen.}
The training constants are simple, standard values, fixed once per
regime before the reported five-seed runs; none is chosen per dataset.
Raw-feature $k$NN uses $k{=}20$, inside the $10<k<50$ range
recommended for neighbor-based detectors
\citep{goldstein2016comparative}. ZEN and FOCUS use $k{=}50$, one
value for both regimes and every dataset.

\subsection{Baseline configurations}
\label{app:baselines}
ADBench benchmarks 14 unsupervised detectors, 13 of them through PyOD.
DAGMM is the exception: ADBench ships its own implementation. We run 12
of the 13 PyOD detectors and leave DAGMM out rather than port a second
implementation of one legacy baseline. PyOD's $k$NN is replaced by raw-feature
$k$NN, the same algorithm scored by the mean distance to its $k{=}20$
nearest neighbors (Appendix~\ref{app:config}) rather than the library
default, the distance to the fifth neighbor. All PyOD detectors run
at library defaults with
contamination${}=0.1$, matching how ADBench itself instantiates them
\citep{han2022adbench}, which likewise runs every detector at its
default hyperparameters rather than tuning on a labeled hold-out set.
Such tuning is not available to a label-free practitioner in any case,
so no method is tuned per dataset (Sec.~\ref{sec:results}). The one
deviation from the defaults is
PCA's component count. At the default the detector keeps every
component, and on three (dataset, seed) pairs the training covariance
is singular, which makes the resulting scores infinite. We therefore
cap the count at 95\% cumulative explained variance, a standard
sklearn idiom, uniformly on every dataset.

The added baselines run as their authors specify. TCCM trains one
model per dataset with its authors' architecture and training
protocol; its per-dataset epoch count and batch size come from the
fixed table their code ships, one entry per dataset, unchanged across
seeds. The per-seed score vectors of its contaminated-regime run were
not retained, only the per-dataset AUROC means, so its contaminated
AUPRC is withheld in Table~\ref{tab:auprc}. It reads
the shared MinMax $[-1,1]$ features like every other method rather
than its own z-scoring default. DTE runs in its categorical variant at
the authors' recommended hyperparameters. The TabPFN pseudo-labeling
control flags the 10\% of training samples a PCA detector scores as
most anomalous, fits TabPFN's classifier on those pseudo-labels, and
scores each test sample by its predicted anomaly probability. The
official unsupervised TabPFN extension (uTabPFN) runs through its
public API; its per-feature density construction is what makes it
infeasible on the three widest datasets (App.~\ref{app:scope}).
The pretrained anomaly detectors of Sec.~\ref{sec:related} (TACTIC,
FoMo-0D, OutFormer, OFA-TAD) are not run: this paper asks what a
general-purpose tabular model gives anomaly detection, so every compared
method starts without anomaly-specific pretraining.

\subsection{Runtime}
\label{app:runtime}
Table~\ref{tab:runtime} reports wall-clock times on a single NVIDIA
A100 GPU in the contaminated regime. The baseline rows come from one
probe on the seed-0 splits; each is the detector's fit plus scoring.
The rows of our methods are stage medians over the 235 runs of the
main evaluation (47 datasets, five seeds) on the same GPU model, with
totals per seed. ZEN's inference has two stages: building the TabPFN
engines and embedding the reference set, then the six feature-subset
embeddings and the $k$NN readout. FOCUS adds its one-time fine-tune,
a training cost, and then runs the same inference on the adapted
model. The probe does not cover TCCM, DTE, uTabPFN, or TabPFN-PL.

\begin{table}[t]
\centering
\caption{Wall-clock time over the 47 contaminated-regime datasets on
one NVIDIA A100-SXM4-80GB. Baselines: fit plus scoring on the seed-0
splits, single run, indicative rather than seed-averaged. Ours: median
per dataset and total per seed of each stage, over the 235 runs of
the official run.}
% Body: final/figures/paper_tab_runtime.tex (build_appendix_v1.py; runtime_probe.py + F102 stage timings).
\label{tab:runtime}
\medskip
\small
\begin{tabular}{lrr}
\toprule
Method & median s\,/\,dataset & total s (47 datasets)\\
\midrule
LODA & 0.03 & 2.7\\
PCA & $<$0.01 & 3.1\\
COPOD & 0.01 & 3.2\\
ECOD & 0.01 & 3.3\\
raw $k$NN & 0.05 & 3.6\\
HBOS & $<$0.01 & 4.5\\
CBLOF & 0.03 & 4.5\\
LOF & 0.08 & 6.5\\
IForest & 0.16 & 7.8\\
OCSVM & 0.70 & 70.6\\
SOD & 1.37 & 95.2\\
DeepSVDD & 6.64 & 325.5\\
COF & 2.77 & 351.5\\
\textbf{ZEN (ours): engines and reference embedding} & 16.6 & 850\\
\textbf{ZEN (ours): six feature-subset embeddings and readout} & 57.3 & 2,904\\
\textbf{FOCUS (ours): fine-tune (training)} & 77.3 & 7,340\\
\textbf{FOCUS (ours): six feature-subset embeddings and readout} & 57.0 & 2,879\\
\bottomrule
\end{tabular}
\end{table}

\section{Negative results}
\label{app:negative}
We record the design alternatives we tried and rejected.
Each was evaluated under the paper's protocol in the contaminated regime
and improved on nothing.
\begin{itemize}
  \item \textbf{LOF scoring instead of $k$NN distance.} Local-density
    normalization loses to plain $k$NN distance, on raw features and on
    our embeddings. Several datasets contain enough exact duplicate samples
    to break LOF's reachability normalization, and a de-duplicated
    variant did not close the gap.
  \item \textbf{Hubness correction.} Rescaling the embedding distances
    by mutual proximity is worse than the plain distance.
  \item \textbf{Whitened and Mahalanobis-style $k$NN.} Not
    significantly better than plain $k$NN in either regime.
  \item \textbf{Distance to the center as the score.} The training
    objective itself, read directly as an anomaly score, is far worse
    than the $k$NN readout in both regimes.
  \item \textbf{Bagging the reference set.} Averaging the readout over
    random subsets of the reference set, instead of over feature
    subsets, brings nothing.
  \item \textbf{Other ways to pool the isolation estimate.} Weighting
    feature subsets by their agreement, pessimistic pooling, calibrating
    the subsets to a common scale, and pooling over two neighborhood sizes
    all fail to improve on the plain mean of Sec.~\ref{sec:zen}.
  \item \textbf{The classifier head.} TabPFN's predicted class
    probability under the arbitrary context labels is at chance as an
    anomaly score, and mixing it into the embedding score hurts.
  \item \textbf{Variants of the fine-tune.} Fine-tuning each feature
    subset separately, fine-tuning through the augmented context, using the
    pooled isolation estimate as the fine-tune's trust weights, and
    contracting each reference sample toward a local center instead of
    one global center bring no gain.
\end{itemize}

\end{document}